\PassOptionsToPackage{table}{xcolor}
\documentclass{article}
\usepackage{iclr2027_conference,times}

\usepackage{amsmath,amsfonts,bm}

\def\eqref#1{equation~\ref{#1}}

\def\1{\bm{1}}

\DeclareMathAlphabet{\mathsfit}{\encodingdefault}{\sfdefault}{m}{sl}
\SetMathAlphabet{\mathsfit}{bold}{\encodingdefault}{\sfdefault}{bx}{n}

\usepackage{hyperref}
\usepackage{url}
\usepackage{graphicx}
\usepackage{caption}
\usepackage{amsmath}
\usepackage{amssymb}
\usepackage{booktabs}
\usepackage{multirow}
\usepackage{array}
\usepackage{tabularx}
\usepackage{pifont}
\usepackage{float}
\usepackage{enumitem}
\usepackage{xcolor}
\usepackage{hhline}
\usepackage{fontawesome5}
\usepackage{pifont}
\definecolor{evibest}{RGB}{200,0,0}
\definecolor{evisecond}{RGB}{70,130,180}
\definecolor{eviheader}{HTML}{DBD9E4}
\definecolor{eviours}{RGB}{254,247,242}
\definecolor{evisubheader}{RGB}{248,248,248}
\definecolor{deltagreen}{RGB}{75,115,95}

\newcommand{\bestres}[1]{\textcolor{evibest}{\textbf{#1}}}
\newcommand{\secondres}[1]{\textcolor{evisecond}{\textbf{#1}}}
\newcommand{\appref}[1]{%
  \hyperref[#1]{Appendix~\ref*{#1}}%
}
\newcommand{\cmark}{\ding{51}}
\newcommand{\xmark}{\ding{55}}

\title{
Pay\hspace{1.2em}More\hspace{1.2em}Attention\hspace{1.2em}to\hspace{1.2em}Text\hspace{1.2em}in
\\
High-Resolution\hspace{1.2em}MLLMs
}
\author{
\textbf{
Zhongkuan Mao$^{1,*}$,
Wenzhuo Zhao$^{2,*}$,
Xianjie Liu$^{2}$,
Yidong Wang$^{3}$,
Zhao Gao$^{2}$,
Ronghao Xian$^{2}$,}\\[-0.05em]
\textbf{
Yao Jiang$^{2}$,
Yi Zhang$^{4}$,
Liangjian Wen$^{5}$,
Keren Fu$^{1,2,\dagger}$}\\[0.45em]
{\footnotesize
\textsuperscript{\rm 1}\,
\textit{National Key Laboratory of Fundamental Science on Synthetic Vision, Sichuan University}
}\\[0.12em]
{\footnotesize
\mbox{\textsuperscript{\rm 2}\,\textit{College of Computer Science, Sichuan University}}
\hspace{2.5em}
\mbox{\textsuperscript{\rm 3}\,\textit{Peking University}}
\hspace{2.5em}
\mbox{\textsuperscript{\rm 4}\,\textit{X-Humanoid}}
}\\[0.12em]
{\footnotesize
\textsuperscript{\rm 5}\,
\textit{Southwest University of Finance and Economics}
\hspace{2.5em}
$^{*}$\,\textit{Equal contribution}
\hspace{2.5em}
$^{\dagger}$\,\textit{Corresponding author}
}
}
\iclrfinalcopy 

\begin{document}

\maketitle
\lhead{}
\renewcommand{\headrulewidth}{0.4pt}

\begin{abstract}
Failures of high-resolution MLLMs are commonly attributed to a visual problem, motivating zooming, cropping, and related visual interventions to recover fine-grained evidence or suppress interference. Yet recent studies suggest that relevant visual evidence is already encoded in intermediate representations, indicating that visual-side improvements alone insufficient. This raises a natural question: does the remaining bottleneck lie in the text that guides visual search? We identify a previously overlooked linguistic bottleneck: questions formulated for answering do not necessarily specify the visual evidence required for localization. To address this mismatch, we introduce EviSpec, a training-free compiler that derives complementary evidence specifications while preserving the original question for final reasoning. We further validate it through matched-control experiments that isolate the roles of evidence specification and localization. With the search budget fixed, structured evidence specifications yield an 11.9\% relative gain over generic requests. With evidence geometry matched, the evidence localized by EviSpec yields a 14.8\% relative gain over random evidence. Together, these controls isolate the benefit of specifying what evidence to seek rather than merely expanding visual access. Across all five MLLMs, EviSpec consistently improves upon the corresponding baseline on each of the three benchmarks, yielding average relative gains of \textbf{10.4\%, 8.8\%, and 12.4\%} on V\textsuperscript{*}Bench, HR-Bench-4K, and HR-Bench-8K, respectively. Beyond high-resolution reasoning, EviSpec also achieves state-of-the-art performance on VQA and hallucination-focused benchmarks.
\end{abstract}

\vspace{-3mm}

\begin{figure}[H]
    \centering
    \captionsetup{skip=5pt}
    \includegraphics[width=1\linewidth]{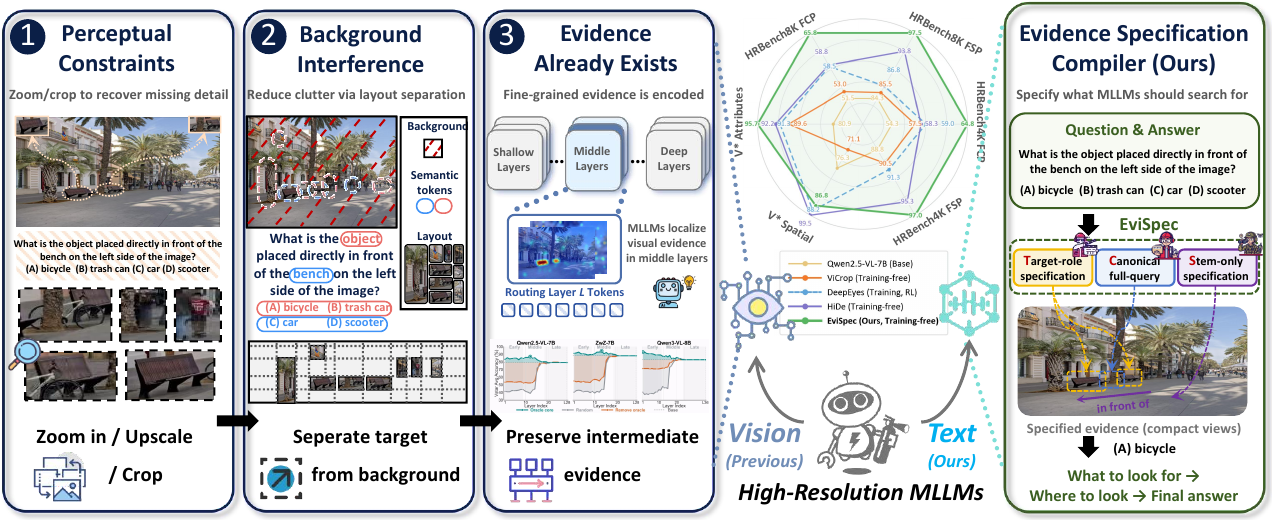}
    \caption{\textbf{Motivation and overview of EviSpec.} Prior work addresses high-resolution failures through \textbf{visual} side, from recovering missing details to suppressing interference and preserving already-encoded evidence. We instead focus on the \textbf{text} guiding visual search. EviSpec consistently outperforms visual baselines and compiles each question into complementary target-role ($T$), canonical full-query ($C$), and stem-only ($S$) evidence specifications. Their ordered combination, $TCS$, specifies what and where to search, while retaining the original question for final reasoning.}
    \label{fig_intro}
\end{figure}

\section{Introduction}
Multimodal large language models (MLLMs) increasingly support native or dynamic-resolution inputs, exposing substantially more visual detail to the language model~\citep{wang2024exploring,jiang2024effectiveness,bai2025qwen25vl,bai2025qwen3vl,liu2025nexus,zhu2025internvl3,zhao2026time,zhao2026samba+}. Yet high-resolution reasoning remains challenging when the answer depends on sparse evidence hidden within a large and cluttered scene, such as a small sign, a distant object, or a relation between spatially separated entities. Benchmarks such as V$^*$Bench and HR-Bench make this challenge explicit by requiring models to recover fine-grained evidence from high-resolution images~\citep{wu2024vstar,wang2025dc2}. As shown in Fig.~\ref{fig_intro}.\ding{182}--\ding{184}, existing work has predominantly addressed the failures on \textbf{the visual side}: zooming into promising regions, selecting crops, invoking visual tools, suppressing irrelevant content, or exploiting internal attention to recover evidence~\citep{khayatkhoei2025mllms,liu2025hide,xie2026havc,zheng2025deepeyes,zhu2026laser}. Despite their methodological differences, these approaches share a premise---high-resolution failure is primarily a problem of \textbf{where and how the model looks}. However, this visual-side view may be incomplete. 

Recent analyses suggest that high-resolution MLLMs can already encode answer-relevant visual content in intermediate representations~\citep{wang2025mllm,mao2026thinking} in Fig~\ref{fig_intro}.\ding{184}. This raises a complementary question that has received far less attention: \emph{what tells the model what visual evidence to look for?} Since visual search is guided by language, we ask whether part of the remaining high-resolution bottleneck lies not in the image itself, but \textbf{in the text used to guide visual search}. Our key observation is that text formulated for answering is not necessarily specified for localization. While answering preserves the user’s decision problem, localization requires explicit cues about the target, its attributes, and relevant relations~\citep{yu2018mattnet,ma2023query}. These requirements can diverge substantially in high-resolution VQA: a question may support correct reasoning while leaving the required visual evidence implicit or ambiguously specified. We call this mismatch the \textbf{evidence-specification gap:} the question states what to decide, but not necessarily what to find.

To address this gap, we introduce \textbf{EviSpec}, a training-free question-to-evidence compiler that provides a dedicated language-side interface for visual search. As illustrated in Fig.~\ref{fig_intro}, EviSpec compiles the original question into \textbf{three complementary evidence specifications}. The \textbf{T}arget-role specification ($T$) identifies the answer-bearing target and relevant reference entities; the \textbf{C}anonical full-query specification ($C$) rewrites the complete question with stable entity expressions; and the option-isolated \textbf{S}tem specification ($S$) applies the same canonicalization after removing candidate answers. Their ordered combination, \textbf{$TCS$}, specifies what evidence to seek and where to search for it. MLLMs then localize these specifications through prompt-calibrated cross-modal attention and adaptive region formation, producing compact evidence views alongside the original image. Rather than changing how the model sees, EviSpec changes the text that guides what it looks for.

\textbf{(1)} Importantly, to isolate this text-side effect from increased visual access, we conduct \textbf{two controlled experiments}. First, with the visual-search budget fixed, structured evidence specifications yield an \textbf{11.9\%} improvement over generic requests, showing that the gain does not come from additional search. Second, with evidence geometry matched, regions localized by EviSpec yield a \textbf{14.8\%} improvement over random evidence, ruling out gains from simply providing an extra crop or favorable region geometry. Together, these controls isolate the benefit of specifying what evidence to seek and retrieving question-relevant content. \textbf{(2)} This text-side intervention produces consistent gains across models and tasks. Across all five evaluated MLLMs, EviSpec improves upon the corresponding baseline on each of the three high-resolution benchmarks, with average relative gains of \textbf{10.4\%, 8.8\%, and 12.4\%} on V$^*$Bench, HR-Bench-4K, and HR-Bench-8K, respectively. The benefit further extends beyond high-resolution reasoning, where EviSpec achieves state-of-the-art (SOTA) performance on additional VQA and hallucination-focused benchmarks. In the end, these results indicate that high-resolution MLLM failures \textbf{cannot be reduced to insufficient visual access}: how the evidence is specified in text is an important part of the multimodal reasoning interface.

Our principal contributions are summarized as follows:

\begin{itemize}[leftmargin=*, labelindent=0pt, labelsep=0.5em]
    \item \textbf{\textit{Problem Reframing.}} We identify an overlooked linguistic bottleneck in high-resolution MLLMs: while prior work primarily improves visual access, \textbf{the text guiding visual search} fails to specify the evidence for localization. We formalize this mismatch as the evidence-specification gap.
    \item \textbf{\textit{Paradigm Reformulation.}} We introduce EviSpec, a training-free question-to-evidence method that derives complementary search specifications while preserving the question for final reasoning. It uses no external detector or model modification, with a single inference hyperparameter.
    \item \textbf{\textit{Controlled and Comprehensive Validation.}} Controlled analyses reveal how evidence specification and question-relevant localization drive the gains beyond additional visual access. Across five frozen MLLMs and eight benchmarks, EviSpec consistently achieves SOTA performance on high-resolution VQA and hallucination-focused evaluations.

\end{itemize}

\begin{figure*}[t]
\centering
\includegraphics[width=\textwidth]{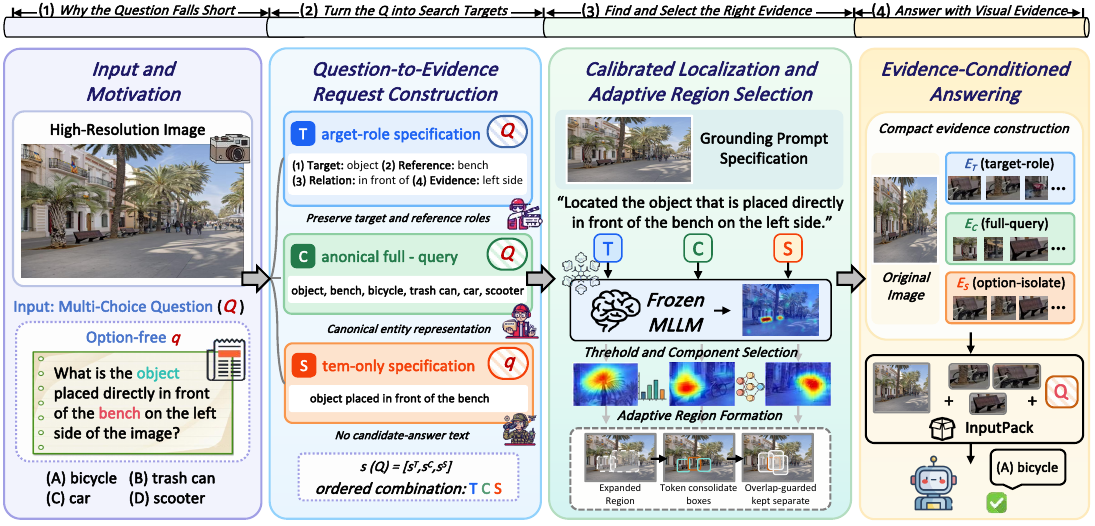}
\caption{Overview of EviSpec: $T$ preserves target roles, $C$ captures the full query, and $S$ isolates answer options. Prompt-calibrated attention extracts compact visual evidence, while adaptive region formation preserves relational context.}
\label{fig:pipeline}
\vspace{-2mm}
\end{figure*}

\section{Related Work}
\paragraph{Multimodal Large Language Models}

MLLMs integrate visual encoders with pretrained language models to support image understanding and visual reasoning~\citep{li2023blip-2,dai2023instructblip,driess2023palm,liu2024improved,zhang2025mm1}. Recent models, including Qwen2.5-VL, Qwen3-VL, and InternVL3, process images at native or dynamic resolutions and preserve fine-grained visual information \citep{bai2023qwenvl,chen2024fargpt4vclosinggap,chen2024internvl,wang2024qwen2vl,bai2025qwen25vl,bai2025qwen3vl,wang2025internvl3.5,zhu2025internvl3}. However, a larger visual input does not ensure that question-relevant evidence is retained, localized, or used during inference. Most advances improve visual encoding, token allocation, or multi-modal alignment~\citep{wen2026infmasking,wen2026dependency}.

\paragraph{High-Resolution Visual Question Answering}

High-resolution visual question answering (HR-VQA) requires models to recover small objects, fine attributes, and relations between spatially separated entities. V* introduced guided visual search and V*Bench, while HR-Bench extended evaluation to 4K and 8K images. Existing methods improve evidence access through regional decomposition, attention-guided cropping, relevance maps, grounding-aware head selection, adaptive localization, evidence routing, or crop policies \citep{you2024ferret,li2024monkey,dong2024internlm,huang2025mini,zhong2025focus,li2026deepscan,lin2026adaptvision,liu2025hide,wei2026zooming,xie2026havc,zheng2025deepeyes,wang2026vgr,zhu2026laser,yuan2026vision}. Question-conditioned grounding and visual question decomposition connect linguistic structure with image regions \citep{shi2018qta,kamath2021mdetr,le2022attentionpriors,li2022grounded,jiang2024diem,peng2024grounding,rasheed2024glamm,liu2024grounding,zhang2024vqd,ma2025clawmachine,wu2025towards,dong2026refadv,dong2026mmtok,huang2026nwa,wu2026postalign}. EviSpec separates evidence specification from localization and answering. It compiles the question into target-role, canonical, and option-isolated localization requests while preserving it for answering.

\section{Methodology}
Fig.~\ref{fig:pipeline} provides an overview of the EviSpec pipeline, including request construction, localization, region formation, and evidence-conditioned answering. Details are provided in the \appref{sec:supplement}.

\subsection{Problem Formulation and Framework Overview} As shown in Fig.~\ref{fig:pipeline} (1) , $I:\Omega_I\rightarrow\mathbb{R}^3$ denote an image on source-pixel domain $\Omega_I$, with visual token grid $\Omega_V=\{1,\ldots,H_V\}\times\{1,\ldots,W_V\}$. A query $Q=(q,\mathcal{O})$ contains a question and an optional set of candidate answers, serialized as $\bar Q=\operatorname{Serialize}(q,\mathcal{O})$; when $\mathcal{O}=\varnothing$, $\bar Q=q$. The compiler, localization, and answer calls, denoted $F_\theta^{\mathrm{comp}}$, $F_\theta^{\mathrm{loc}}$, and $F_\theta^{\mathrm{ans}}$, share frozen parameters but use distinct fixed prompts. The evidence constructor $G_\theta$ invokes the compiler and localizer to produce ordered views $\mathbf{E}=G_\theta(I,Q)$. Final inference retains the original image and query:
\begin{equation}
\hat a=
F_\theta^{\mathrm{ans}}\!\left(
\operatorname{InputPack}\!\left([I]\Vert G_\theta(I,Q)\right),Q
\right).
\label{eq:evidence-stage}
\end{equation}
Evidence construction uses neither gold answers, annotation boxes, nor benchmark categories. Only the final call performs task-specific answer reasoning; no parameters are trained.

\subsection{Question-to-Evidence Request Construction}
As shown in Fig.~\ref{fig:pipeline} (2) , the option-free stem $q_0$ is the prefix of $\bar Q$ before its first option marker, or $q$ when no options are present. Fixed compiler prompts produce the ordered specifications
\begin{equation}
\widetilde{\mathbf{s}}(\bar Q)
=[s^T,s^C,s^S]
=\bigl[\operatorname{Role}(\bar Q),
       \operatorname{Can}(\bar Q),
       \operatorname{Can}(q_0)\bigr].
\label{eq:specifications}
\end{equation}
For direct questions, $\operatorname{Role}$ returns one answer-bearing target phrase with its identifying context. For relation questions, it returns comma-delimited subject and reference fields in question order and records relation mode. Canonicalization rewrites entities into head-first phrases: $C$ uses the complete query, whereas $S$ applies the same contract to the option-free stem. $\mathbf{s}(\bar Q)$ removes empty parsed requests and exact duplicates, preserving $T/C/S$ order. Let $\mathcal{R}(Q)\subseteq{T,C,S}$ denote the surviving branch labels. Each retained request has the representation $s^r=(u_{r,1},\ldots,u_{r,K_r};m_r)$ for $r\in\mathcal{R}(Q)$. Here, $u_{r,j}$ is an entity field and $m_r$ an optional direct or relation-oriented mode. Token group $\mathcal{T}*{r,j}$ contains signal tokens aligned with $u*{r,j}$. All branches share the localization procedure below and terminate before final answering. Details are provided in \appref{sec:app-prompt-controls}.

\subsection{Calibrated Localization and Adaptive Region Selection
}
As shown in Fig.~\ref{fig:pipeline} (3) , each retained request is inserted into a shared grounding prompt and localized on the original image by $F_\theta^{\mathrm{loc}}$. $A_{r,t}^{(h)}(p)$ denote the attention of head $h$ from token $t$ to visual location $p$. With $\mathcal{H}_F$ indexing all attention heads, the response is
\begin{equation}
A_{r,t}(p)=
\frac{1}{|\mathcal{H}_F|}
\sum_{h\in\mathcal{H}_F}A_{r,t}^{(h)}(p).
\label{eq:attn}
\end{equation}
Each nonconstant map is min--max normalized to place its responses on a common scale:
\begin{equation}
\operatorname{N}(X)(p)=
\frac{X(p)-\min_{z\in\Omega_V}X(z)}
{\max_{z\in\Omega_V}X(z)-\min_{z\in\Omega_V}X(z)}.
\label{eq:normalization}
\end{equation}
Constant maps are assigned zero, and we write $\bar A_{r,t}=\operatorname{N}(A_{r,t})$. Instruction-prefix tokens can activate salient regions~\cite{chen2026accelerating}; their token set $\mathcal{P}_r$ defines a pointwise median prior:
\begin{equation}
B_r(p)=
\operatorname{median}_{u\in\mathcal{P}_r}
\operatorname{N}(A_{r,u})(p).
\label{eq:prompt-prior}
\end{equation}
Prefix tokens end at the grounding instruction's colon; chat-template suffix tokens are excluded. Each signal-token response is calibrated separately:
\begin{equation}
R_{r,t}(p)=
\operatorname{N}\!\left([\bar A_{r,t}-B_r]_+\right)(p).
\label{eq:calibrated-response}
\end{equation}
Here, $[\cdot]+$ clips negative values to zero. If prior removal produces an identically zero map, we retain the original $\bar A{r,t}$. Signal-token maps are kept separate to avoid diluting informative entity- or attribute-specific responses through averaging with unrelated tokens. For each calibrated response map $R_{r,t}$, we independently compute a threshold
$\tau_{r,t}=\operatorname{Yen}(R_{r,t})$ using Yen's criterion~\citep{yen1995threshold}. Thresholding $R_{r,t}$ at $\tau_{r,t}$ yields the foreground support
${p\in\Omega_V:R_{r,t}(p)\geq\tau_{r,t}}$, which is partitioned into eight-connected components collected in $\mathcal{C}{r,t}^{+}$. We then select as the seed the component with the largest total response mass:
\begin{equation}
C{r,t}^{}=
\operatorname{arg,max}{C\in\mathcal{C}{r,t}^{+}}
\sum_{p\in C}R_{r,t}(p).
\label{eq}
\end{equation}
Equal-mass ties are resolved in row-major order. Expansion is enabled for requests with multiple entity fields ($K_r>1$), as in relation-mode $T$ requests. Let $V_{r,t}^{-}$ denote response values below $\tau_{r,t}$; when nonempty, $\tau_{r,t}^{-}=\operatorname{Yen}(V_{r,t}^{-})$ defines
\begin{equation}
C_{r,t}^{-}=
\operatorname{CC}_8\!\left(
\{p\in\Omega_V:R_{r,t}(p)\geq\tau_{r,t}^{-}\};
C_{r,t}^{*}
\right).
\label{eq:relational}
\end{equation}
$\operatorname{CC}_8(\cdot;C_{r,t}^{*})$ denotes the eight-connected component containing the original seed. If expansion is disabled or $V_{r,t}^{-}$ is empty, the seed is retained. Selected components are mapped to source-pixel boxes in $\Omega_I$, and token groups associate these boxes with entities. An enclosing entity box replaces its token boxes only if it has no positive-area overlap with another entity's boxes. Otherwise, the original token boxes are preserved to avoid absorbing a reference entity. Detailed in \appref{sec:app-complete-operators}.

\subsection{Evidence-Conditioned Answering}
As shown in Fig.~\ref{fig:pipeline} (4) , for each surviving request, $\mathcal{B}_r$ contain its retained boxes in row-major order. Their source pixels are assembled on a transparent canvas:
\begin{equation}
E_r=
\operatorname{Assemble}\!\left(
[\Pi(I,b)]_{b\in\mathcal{B}_r}
\right),
\qquad r\in\mathcal{R}(Q).
\label{eq:evidence-view}
\end{equation}
$\Pi(I,b)$ copies the pixels inside $b$; assembly removes empty gaps while preserving the retained pixels and region order. Exact duplicate views are removed in $T/C/S$ order to form $\mathbf{E}$. The model's native multi-image interface receives
\begin{equation}
\mathbf{X}_E=
\operatorname{InputPack}\!\left(
[I]\Vert
\operatorname{UniqueView}\!\left([E_r]_{r\in\mathcal{R}(Q)}\right)
\right).
\label{eq:unique-evidence}
\end{equation}
The original image supplies the global spatial reference, while $\hat a=F_\theta^{\mathrm{ans}}(\mathbf{X}_E,Q)$ preserves the original query, task-specific answer prompt, and decoding rule. All benchmarks share these operators, without external detectors, learned fusion weights, or benchmark-conditioned branches.

\begin{figure*}[t]
    \centering
    \captionsetup{skip=5pt}
    \includegraphics[width=\textwidth]{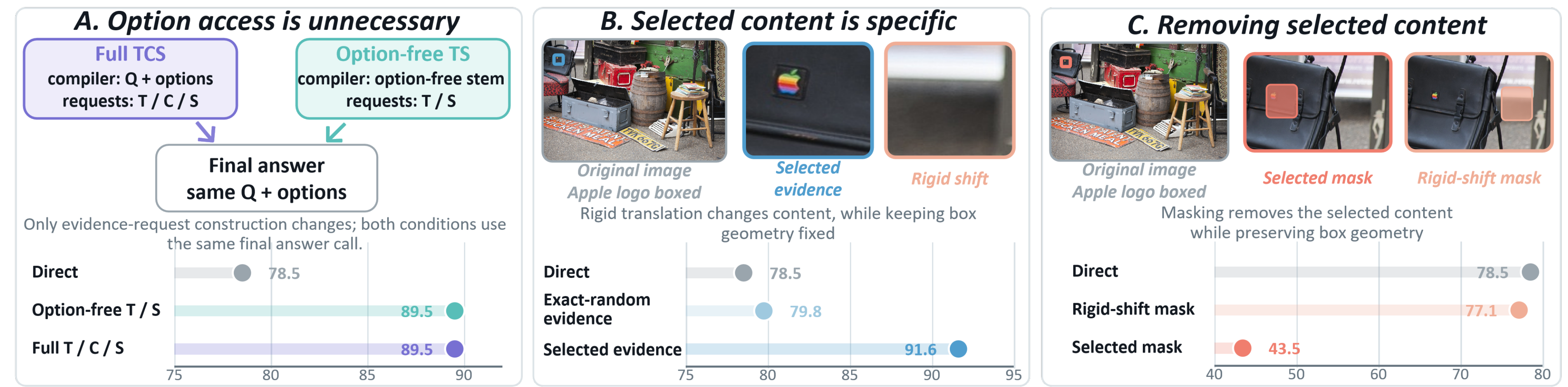}
    \caption{
    \textbf{Matched controls on V$^*$Bench with Qwen2.5-VL-7B.}
    (A) Option-free $TS$ matches full-input $TCS$ while using the same final answer call.
    (B) Selected evidence outperforms exact-geometry random evidence (91.6\% vs. 79.8\%).
    (C) Masking selected regions causes a substantially larger accuracy drop than masking rigidly shifted regions (43.5\% vs. 77.1\%).
    }
    \label{fig:ablation_analysis2}
    \vspace{-3mm}
\end{figure*}

\section{Analyses of Evidence Specification}
\label{sec:analysis}

To isolate the effect of structured requests from increased visual access, we conduct \textbf{two controlled experiments}: one on localization and one on the utility of localized evidence for answering. These analyses control for request multiplicity, spatial coverage, and access to answer choices. All experiments use Qwen2.5-VL-7B on V$^*$Bench, with protocols detailed in \appref{sec:app-matched-controls} and \appref{sec:app-selected-content}.

\subsection{Disentangling Request Structure from Multiplicity}
\label{sec:request-budget}

We first explore whether EviSpec benefits from generating more localization requests or from expressing the evidence need more effectively. Generic request generation and $TCS$ each receive three generation attempts before deduplication and evidence assembly.$TCS$ achieves an accuracy of 88.5\%, compared with 79.1\%, representing an absolute improvement of 9.4 percentage points and a relative gain of 11.9\%. Yet $TCS$ retains fewer distinct non-empty requests after deduplication, averaging 1.9 rather than 2.6. The combination of higher accuracy and fewer retained requests distinguishes request quality from request quantity. In this control, the improvement cannot be attributed to a larger retained request set. Instead, the result supports explicitly organizing the information supplied to the localizer, rather than treating additional formulations as inherently useful search instructions. This distinction motivates EviSpec's compiler interface: the objective is to specify the required visual evidence, not simply to produce more versions of the question.

\subsection{Disentangling Request Roles from Spatial Extent}
\label{sec:request-localization}

\begin{figure*}[t]
  \centering
  \captionsetup{skip=5pt}
  \includegraphics[width=\textwidth]
  {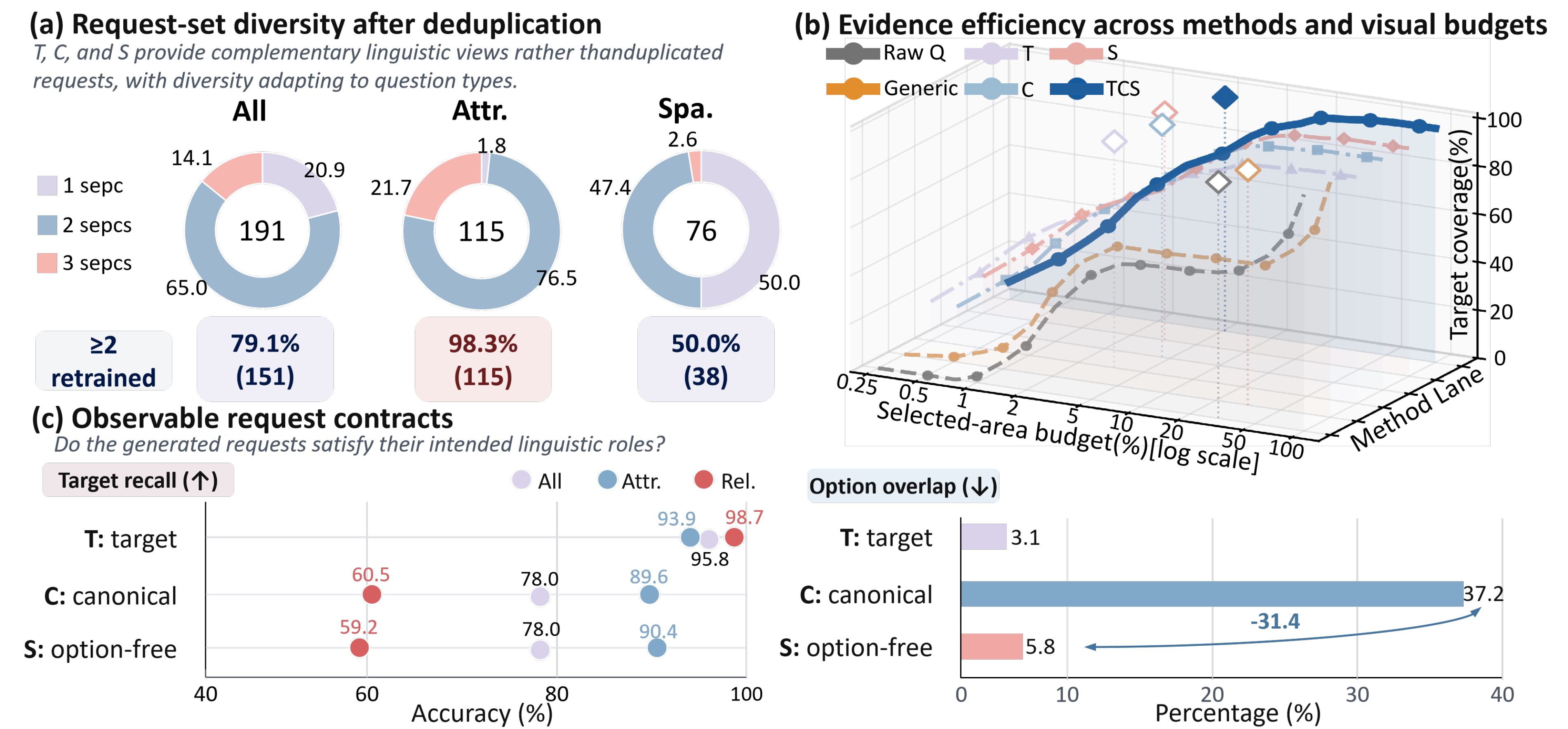}
  \caption{
    Analysis of evidence specifications:
    (a) diversity across $T$, $C$, and $S$,
    (b) evidence efficiency under varying visual budgets, and
    (c) semantic differences in their intended localization roles.
  }
  \label{fig:ablation_analysis}
\end{figure*}

The request profiles in Fig.~\ref{fig:ablation_analysis} (a,c) help interpret what this organization changes. The branches exhibit different degrees of target preservation and answer-option overlap, with different subsets retained for attribute and spatial questions. These properties reflect different treatments of the same evidence need: $T$ foregrounds target roles, $C$ canonicalizes the complete query, and $S$ isolates the question stem. Exact-string deduplication removes duplicate outputs while preserving distinct formulations, yielding a question-dependent request set rather than a fixed three-request expansion. The spatial consequences are examined through target coverage under selected-area budgets in Fig.~\ref{fig:ablation_analysis} (b). Coverage alone can reward broad selections that include the target with substantial surrounding content. The area constraint makes this distinction explicit by asking whether target coverage is retained without exceeding the permitted spatial extent. At intermediate budgets, $TCS$ achieves higher coverage than raw questions and generic requests. Its advantage concerns the concentration of target-relevant content within compact selections, rather than coverage obtained through unrestricted enlargement. The request profiles and coverage curves provide complementary observations: the former characterize the linguistic inputs, while the latter reveal differences in spatial selectivity.

\begin{table*}[t]
  \centering
  \caption{Answer accuracy (\%) on V$^*$Bench and HR-Bench. 
  \bestres{Red} and \secondres{blue} entries mark the best and second-best 
  distinct results within each open-source base-model block; 
  $\Delta$ is EviSpec minus Base.}
  \label{tab:main_results}

  {
  \footnotesize
  \renewcommand{\arraystretch}{1.04}
  \setlength{\extrarowheight}{0pt}
  \setlength{\tabcolsep}{2.0pt}
  \setlength{\arrayrulewidth}{0.4pt}

  \renewcommand{\tabularxcolumn}[1]{m{#1}}

  \begin{tabularx}{\textwidth}{
    >{\centering\arraybackslash}m{2.20cm}|
    >{\centering\arraybackslash}m{2.75cm}|
    >{\centering\arraybackslash}m{0.95cm}|
    *{3}{>{\centering\arraybackslash}X}|
    *{3}{>{\centering\arraybackslash}X}|
    *{3}{>{\centering\arraybackslash}X}
  }
    \noalign{\hrule height 1pt}
    \rowcolor{eviheader}
    \multicolumn{12}{c}{\textbf{\faEye~~High-Resolution Visual Question Answering}} \\
    \hline

    \rowcolor{eviheader}
    \rule{0pt}{2.8ex}
    &
    &
    & \multicolumn{3}{c|}{\textbf{V$^*$}}
    & \multicolumn{3}{c|}{\textbf{HRBench-4K}}
    & \multicolumn{3}{c}{\textbf{HRBench-8K}}
    \\

    \rowcolor{eviheader}
    \multirow{-2}{*}{\textbf{Method}}
    & \multirow{-2}{*}{\textbf{Base Model}}
    & \multirow{-2}{*}{\shortstack[c]{\textbf{Train}\\\textbf{Free}}}
    & \textbf{Attr}
    & \textbf{Spa.}
    & \textbf{Avg}
    & \textbf{FSP}
    & \textbf{FCP}
    & \textbf{Avg}
    & \textbf{FSP}
    & \textbf{FCP}
    & \textbf{Avg}
    \\

    \hline

    --
    & GPT-4o
    & --
    & -- & -- & 66.0
    & 70.0 & 48.0 & 59.0
    & 62.0 & 49.0 & 55.5
    \\

    \hline

    Base
    &
    & --
    & 80.9 & 61.8 & 73.3
    & 83.0 & 50.3 & 66.6
    & 79.8 & 45.0 & 62.4
    \\

    ViCrop
    &
    & \cmark
    & 81.7 & \secondres{65.8} & 75.4
    & 86.3 & 49.8 & 68.0
    & 80.3 & 45.5 & 62.9
    \\

    HiDe
    &
    & \cmark
    & \secondres{85.2} & \bestres{71.1} & \secondres{79.6}
    & \secondres{87.8} & \secondres{51.5} & \secondres{69.6}
    & \secondres{88.3} & \secondres{51.3} & \secondres{69.8}
    \\

    \rowcolor{eviours}
    \textbf{EviSpec}
    &
    & \cmark
    & \bestres{90.4} & \bestres{71.1} & \bestres{82.7}
    & \bestres{92.3} & \bestres{55.3} & \bestres{73.8}
    & \bestres{89.3} & \bestres{53.8} & \bestres{71.5}
    \\

    $\Delta$ (vs. Base)
    & \multirow{-5}{*}{Qwen2.5-VL 3B}
    & --
    & +9.5 & +9.3 & +9.4
    & +9.3 & +5.0 & +7.2
    & +9.5 & +8.8 & +9.1
    \\

    \hline

    Base
    &
    & --
    & 80.9 & 76.3 & 79.1
    & 88.8 & 54.3 & 71.5
    & 84.3 & 51.5 & 67.9
    \\

    ViCrop
    &
    & \cmark
    & 89.6 & 71.1 & 82.2
    & 90.5 & 57.5 & 74.0
    & 85.5 & 53.0 & 69.3
    \\

    DeepEyes
    &
    & \xmark
    & 91.3 & \secondres{88.2} & 90.1
    & 91.3 & 59.0 & 75.1
    & 86.8 & 58.5 & 72.6
    \\

    HiDe
    &
    & \cmark
    & 92.2 & \bestres{89.5} & \secondres{91.1}
    & \secondres{95.3} & 58.3 & \secondres{76.8}
    & \secondres{93.8} & \secondres{58.8} & \secondres{76.3}
    \\

    DeepScan
    &
    & \cmark
    & \secondres{93.0} & 86.8 & 90.6
    & 90.1 & \secondres{59.7} & 75.0
    & 87.2 & 57.6 & 72.4
    \\

    \rowcolor{eviours}
    \textbf{EviSpec}
    &
    & \cmark
    & \bestres{95.7} & 86.8 & \bestres{92.1}
    & \bestres{97.0} & \bestres{64.8} & \bestres{80.9}
    & \bestres{97.5} & \bestres{65.8} & \bestres{81.6}
    \\

    $\Delta$ (vs. Base)
    & \multirow{-7}{*}{Qwen2.5-VL 7B}
    & --
    & +14.8 & +10.5 & +13.0
    & +8.2 & +10.5 & +9.4
    & +13.2 & +14.3 & +13.7
    \\

    \hline

    Base
    &
    & --
    & 87.8 & 88.1 & 87.9
    & 89.8 & 58.0 & 73.9
    & 84.5 & 56.3 & 70.4
    \\

    ViCrop
    &
    & \cmark
    & 88.7 & 85.5 & 87.4
    & 90.5 & \secondres{59.5} & 74.6
    & 88.8 & 54.3 & 71.5
    \\

    HiDe
    &
    & \cmark
    & \bestres{91.3} & \secondres{89.5} & \bestres{90.6}
    & \secondres{92.0} & \bestres{60.0} & \secondres{76.2}
    & \secondres{90.5} & \secondres{57.5} & \secondres{74.0}
    \\

    \rowcolor{eviours}
    \textbf{EviSpec}
    &
    & \cmark
    & \secondres{89.6} & \bestres{90.8} & \secondres{90.1}
    & \bestres{93.8} & \bestres{60.0} & \bestres{76.6}
    & \bestres{94.5} & \bestres{59.8} & \bestres{77.1}
    \\

    $\Delta$ (vs. Base)
    & \multirow{-5}{*}{Qwen2.5-VL 32B}
    & --
    & +1.8 & +2.7 & +2.2
    & +4.0 & +1.5 & +2.7
    & +10.0 & +3.5 & +6.7
    \\

    \hline

    Base
    &
    & --
    & 87.0 & 78.9 & 83.8
    & 90.2 & 63.7 & 77.0
    & 84.0 & 62.3 & 73.1
    \\

    ViCrop
    &
    & \cmark
    & 89.6 & 75.0 & 83.8
    & 91.8 & 64.3 & 78.0
    & 88.0 & 63.3 & 75.6
    \\

    HiDe
    &
    & \cmark
    & 88.7 & 84.2 & 86.9
    & \secondres{93.8} & \secondres{68.5} & \secondres{81.1}
    & \bestres{94.8} & \secondres{64.0} & \secondres{79.4}
    \\

    DeepScan
    &
    & \cmark
    & \secondres{92.2} & \bestres{89.5} & \secondres{91.1}
    & 92.5 & 64.5 & 78.5
    & 90.0 & 62.3 & 76.1
    \\

    \rowcolor{eviours}
    \textbf{EviSpec}
    &
    & \cmark
    & \bestres{94.8} & \secondres{86.8} & \bestres{91.6}
    & \bestres{94.8} & \bestres{70.3} & \bestres{82.5}
    & \secondres{94.0} & \bestres{66.8} & \bestres{80.4}
    \\

    $\Delta$ (vs. Base)
    & \multirow{-5}{*}{Qwen3-VL 8B}
    & --
    & +7.8 & +7.9 & +7.8
    & +4.6 & +6.6 & +5.5
    & +10.0 & +4.5 & +7.3
    \\

    \hline

    Base
    &
    & --
    & 81.7 & \secondres{78.9} & 80.6
    & 82.8 & \secondres{58.8} & 70.8
    & 80.0 & \bestres{59.8} & 69.9
    \\

    ViCrop
    &
    & \cmark
    & 88.7 & 75.0 & 83.3
    & 88.0 & 57.0 & 72.5
    & 82.8 & \secondres{54.8} & 68.8
    \\

    HiDe
    &
    & \cmark
    & \bestres{92.2} & \bestres{88.2} & \bestres{90.6}
    & \secondres{91.8} & 58.3 & \secondres{75.0}
    & \secondres{86.3} & 53.8 & \secondres{70.0}
    \\

    \rowcolor{eviours}
    \textbf{EviSpec}
    &
    & \cmark
    & \secondres{90.4} & \bestres{88.2} & \secondres{89.5}
    & \bestres{93.8} & \bestres{61.0} & \bestres{77.4}
    & \bestres{90.8} & \bestres{59.8} & \bestres{75.3}
    \\

    $\Delta$ (vs. Base)
    & \multirow{-5}{*}{InternVL3 8B}
    & --
    & +8.7 & +9.3 & +8.9
    & +11.0 & +2.2 & +6.6
    & +10.8 & +0.0 & +5.4
    \\

    \hline
  \end{tabularx}
  }

\end{table*}

\subsection{Disentangling Selected Content from Visual Access}
\label{sec:evidence-relevance}

Compact localization is useful only if it preserves information needed for the final answer. \textbf{First,} we examine whether constructing such evidence requires access to the candidate answers. In Fig.~\ref{fig:ablation_analysis2} (A), option-free $TS$ and full-input $TCS$ both achieve 89.5\% accuracy, while the final answer call receives the same question and options in both settings. Thus, the question stem alone is sufficient to retain the observed aggregate accuracy in this control, supporting a separation between the information used to guide visual search and the complete input used for answer selection.

\textbf{Second,} we test whether the benefit comes from the selected content rather than merely from receiving additional evidence views. A shared rigid translation preserves box count, dimensions, overlap, union area, and packed-view dimensions while altering the sampled image content. As shown in Fig.~\ref{fig:ablation_analysis2} (B), selected evidence achieves 91.6\% accuracy, compared with 79.8\% for geometry-matched random evidence, corresponding to an 11.8-point absolute gain and a 14.8\% relative improvement. With geometry held fixed, this difference supports the value of selecting question-relevant content.

\textbf{Third,} we perform the complementary intervention by removing the selected content. In Fig.~\ref{fig:ablation_analysis2} (C), accuracy drops to 43.5\% when selected regions are masked, compared with 77.1\% for geometrically matched shifted regions. Thus, removing selected regions is substantially more disruptive than removing matched alternatives. Together, the provision and removal interventions show from opposite directions that the localized regions contain information relevant to the downstream answer.

\begin{figure*}[t]
\begin{minipage}{\textwidth}
\captionsetup{type=table}
  \centering
  \caption{
  Additional-task accuracy (\%) under
  a shared protocol across benchmarks. Within each model block, \bestres{red} and
  \secondres{blue} bold values indicate the best and second-best distinct results.
  }
  \label{tab:additional_results}
  {
  \footnotesize
  \renewcommand{\arraystretch}{1.04}
  \setlength{\tabcolsep}{0.75pt}
  \setlength{\arrayrulewidth}{0.35pt}

  \renewcommand{\tabularxcolumn}[1]{m{#1}}

  \begin{tabularx}{\textwidth}{
    >{\raggedright\arraybackslash}m{1.68cm}|
    *{4}{>{\centering\arraybackslash}X}|
    *{3}{>{\centering\arraybackslash}X}|
    *{3}{>{\centering\arraybackslash}X}|
    *{3}{>{\centering\arraybackslash}X}|
    >{\centering\arraybackslash}m{1.02cm}
  }

    \noalign{\hrule height 1pt}
    \rowcolor{eviheader}
    \multicolumn{15}{c}{
      \textbf{\faIcon{exchange-alt}~~Transfer Across Visual Tasks}
    } \\
    \hline

    \rowcolor{eviheader}
    \rule{0pt}{2.55ex}
    &
    \multicolumn{4}{c|}{\textbf{POPE}}
    &
    \multicolumn{3}{c|}{\hspace*{0.6mm}\textbf{ZoomBench}}
    &
    \multicolumn{3}{c|}{\textbf{MME-RW-Lite}}
    &
    \multicolumn{3}{c|}{\textbf{TreeBench}}
    &
    \multicolumn{1}{>{\columncolor{eviheader}}c}{\textbf{DocVQA}}
    \\

    \rowcolor{eviheader}
    \multirow{-2}{*}{\textbf{Method}}
    & \textbf{Adv}
    & \textbf{Pop}
    & \textbf{Ran}
    & \textbf{Avg}
    & \hspace*{0.3mm}\textbf{MCQ}
    & \hspace*{0.3mm}\makebox[\linewidth][c]{\textbf{Blank}}
    & \hspace*{0.6mm}\textbf{Avg}
    & \textbf{Per}
    & \textbf{Rea}
    & \textbf{Avg}
    & \textbf{Per}
    & \textbf{Rea}
    & \textbf{Avg}
    & \multicolumn{1}{>{\columncolor{eviheader}}c}{\textbf{Acc}}
    \\
    \hline

    \rowcolor{evisubheader}
    \multicolumn{15}{c}{
      \rule{0pt}{2.15ex}
      \textbf{\textit{Qwen2.5-VL-7B-Instruct}}
    }
    \\

    Base
    & 84.0 & 84.4 & 85.1 & 84.5
    & 49.9 & \bestres{37.1} & 46.5
    & 51.6 & 39.3 & 46.8
    & 43.6 & 33.2 & 37.0
    & 81.7
    \\

    ViCrop
    & 84.9 & 85.1 & 85.7 & 85.2
    & 41.2 & 26.8 & 37.4
    & 55.6 & 41.6 & 50.1
    & 40.3 & \secondres{34.4} & 36.5
    & 81.8
    \\

    HiDe
    & 85.1
    & 85.4
    & 86.3
    & 85.6
    & \secondres{58.1}
    & \secondres{36.2}
    & \secondres{52.3}
    & \bestres{57.8}
    & \secondres{42.1}
    & \secondres{51.7}
    & \secondres{49.0}
    & 34.0
    & \secondres{39.5}
    & 81.8
    \\
    DeepScan
    & \secondres{85.4} & \secondres{86.0} & \secondres{86.8} & \secondres{86.1}
    & 53.5 & 34.4 & 48.4
    & 49.0 & 37.2 & 44.4
    & 47.0 & 32.0 & 37.5
    & \secondres{82.1}
    \\

    \rowcolor{eviours}
    \textbf{EviSpec}
    & \bestres{86.4}
    & \bestres{87.3}
    & \bestres{89.2}
    & \bestres{87.6}
    & \bestres{61.8}
    & \bestres{37.1}
    & \bestres{55.3}
    & \secondres{56.9}
    & \bestres{45.2}
    & \bestres{52.3}
    & \bestres{50.3}
    & \bestres{34.8}
    & \bestres{40.5}
    & \bestres{82.7}
    \\

    $\Delta$ (vs. Base)
    & +2.4 & +2.9 & +4.1 & +3.1
    & +11.9 & +0.0 & +8.8
    & +5.3 & +5.9 & +5.5
    & +6.7 & +1.6 & +3.5
    & +1.0
    \\

    \hline

    \rowcolor{evisubheader}
    \multicolumn{15}{c}{
      \rule{0pt}{2.15ex}
      \textbf{\textit{Qwen3-VL-8B-Instruct}}
    }
    \\

    Base
    & 87.2
    & \secondres{88.6}
    & 90.9
    & 88.9
    & 46.4
    & 34.4
    & 43.2
    & \secondres{55.9}
    & 43.6
    & 51.1
    & 56.4
    & \secondres{37.8}
    & \secondres{44.7}
    & 88.7
    \\

    HiDe
    & \secondres{87.5}
    & \secondres{88.6}
    & \secondres{91.8}
    & \secondres{89.3}
    & \secondres{57.2}
    & 34.8
    & 51.2
    & 55.3
    & \secondres{45.9}
    & \secondres{51.6}
    & \secondres{57.7}
    & 36.7
    & 44.4
    & 89.0
    \\

    DeepScan
    & 86.7 & 88.5 & 91.2 & 88.7
    & 56.0 & \secondres{41.5} & \secondres{52.2}
    & 52.8 & 40.1 & 47.8
    & 53.7 & 36.3 & 42.7
    & \secondres{89.7}
    \\

    \rowcolor{eviours}
    \textbf{EviSpec}
    & \bestres{87.8}
    & \bestres{88.7}
    & \bestres{92.1}
    & \bestres{89.6}
    & \bestres{62.0}
    & \bestres{45.1}
    & \bestres{57.5}
    & \bestres{58.3}
    & \bestres{47.9}
    & \bestres{54.2}
    & \bestres{58.4}
    & \bestres{39.8}
    & \bestres{46.7}
    & \bestres{90.0}
    \\

    $\Delta$ (vs. Base)
    & +0.6 & +0.1 & +1.3 & +0.7
    & +15.6 & +10.7 & +14.3
    & +2.4 & +4.3 & +3.1
    & +2.0 & +2.0 & +2.0
    & +1.3
    \\

    \noalign{\hrule height 1pt}

  \end{tabularx}
  }

\end{minipage}
\par\vspace{\floatsep}
\begin{minipage}{\textwidth}
\captionsetup{type=table}
  \centering
  \caption{
  Controlled request ablation on Qwen2.5-VL-7B.
  Entries report answer accuracy (\%). \bestres{Red} and \secondres{blue} bold values indicate
  the best and second-best distinct results, respectively.
  }
  \label{tab:ablation}
  {
  \footnotesize
  \renewcommand{\arraystretch}{1.04}
  \setlength{\tabcolsep}{1.10pt}
  \setlength{\arrayrulewidth}{0.35pt}

  \renewcommand{\tabularxcolumn}[1]{m{#1}}

  \begin{tabularx}{\textwidth}{
    >{\raggedright\arraybackslash}m{1.55cm}|
    *{3}{>{\centering\arraybackslash}X}|
    *{3}{>{\centering\arraybackslash}X}|
    *{3}{>{\centering\arraybackslash}X}
  }

    \noalign{\hrule height 1pt}

    \rowcolor{eviheader}
    \rule{0pt}{2.55ex}
    &
    \multicolumn{3}{c|}{\textbf{V$^*$}}
    &
    \multicolumn{3}{c|}{\textbf{HRBench-4K}}
    &
    \multicolumn{3}{c}{\textbf{HRBench-8K}}
    \\

    \rowcolor{eviheader}
    \multirow{-2}{*}{\textbf{Config.}}
    &
    \textbf{Attr}
    & \textbf{Spa.}
    & \textbf{Avg}
    &
    \textbf{FSP}
    & \textbf{FCP}
    & \textbf{Avg}
    &
    \textbf{FSP}
    & \textbf{FCP}
    & \textbf{Avg}
    \\

    \hline

    $T$
    & 93.0
    & 81.6
    & 88.5
    & 94.0
    & 62.5
    & 78.3
    & 94.8
    & \secondres{62.5}
    & \secondres{78.6}
    \\

    $C$
    & 91.3
    & 82.9
    & 88.0
    & 95.0
    & 57.0
    & 76.0
    & 92.5
    & 59.5
    & 76.0
    \\

    $S$
    & 91.3
    & 82.9
    & 88.0
    & 94.5
    & 60.8
    & 77.6
    & 93.5
    & 62.3
    & 77.9
    \\


    $TC$
    & 93.0
    & 82.9
    & 89.0
    & 95.0
    & \secondres{63.0}
    & \secondres{79.0}
    & 94.3
    & 62.0
    & 78.1
    \\

    $TS$
    & 93.0
    & \secondres{84.2}
    & 89.5
    & 95.3
    & 62.8
    & \secondres{79.0}
    & \secondres{95.8}
    & 61.0
    & 78.4
    \\

    $CS$
    & \secondres{93.9}
    & \secondres{84.2}
    & \secondres{90.1}
    & \secondres{95.8}
    & 60.5
    & 78.1
    & 93.3
    & 61.8
    & 77.5
    \\

    \hline

    \rowcolor{eviours}
    \textbf{$TCS$}
    & \bestres{95.7}
    & \bestres{86.8}
    & \bestres{92.1}
    & \bestres{97.0}
    & \bestres{64.8}
    & \bestres{80.9}
    & \bestres{97.5}
    & \bestres{65.8}
    & \bestres{81.6}
    \\

    \noalign{\hrule height 1pt}

  \end{tabularx}
  }
\end{minipage}
\end{figure*}

\section{Experiments}
\label{sec:experiments}
\subsection{Experimental Setup}
\paragraph{Benchmarks and metrics}
We evaluate V*Bench, HR-Bench-4K and HR-Bench-8K \citep{wu2024vstar,wang2025dc2}. V*Bench contains 191 questions, including 115 direct-attribute and 76 relative-position questions. Each HR-Bench split contains 800 questions divided evenly between fine-grained single-region perception (FSP) and cross-region perception (FCP). We report exact-answer accuracy after normalizing each response to one option letter. To examine transfer beyond these core benchmarks, we further evaluate POPE, ZoomBench, MME-RealWorld-Lite, TreeBench, and DocVQA. \citep{mathew2021docvqa,li2023pope,zhang2025mmerealworld,wang2026treebench,wei2026zooming}. POPE is reported on its adversarial, popular and random splits; ZoomBench separates multiple-choice (MCQ) and open-ended blank questions; MME-RealWorld-Lite and TreeBench report perception and reasoning accuracy; and DocVQA reports answer accuracy.

\paragraph{Models and inference}
The completed evaluation covers five frozen MLLMs: Qwen2.5-VL-3B, Qwen2.5-VL-7B, Qwen2.5-VL-32B \citep{bai2025qwen25vl}, Qwen3-VL-8B \citep{bai2025qwen3vl}, and InternVL3-8B \citep{zhu2025internvl3}. The main-table Qwen executions and the branch ablation use a maximum pixel budget of $16{,}384\times28^2$. We enable FlashAttention-2~\cite{dao2024flashattention} for efficient attention computation. EviSpec requires no learned parameters. Compiler prompts, attention processing, region formation and evidence-conditioned answering are shared across datasets. Complete implementation settings and supplementary evaluations are reported in \appref{sec:app-implementation-details}.

\begin{figure*}[t]
    \centering
    \captionsetup{skip=5pt}
    \includegraphics[width=\textwidth]{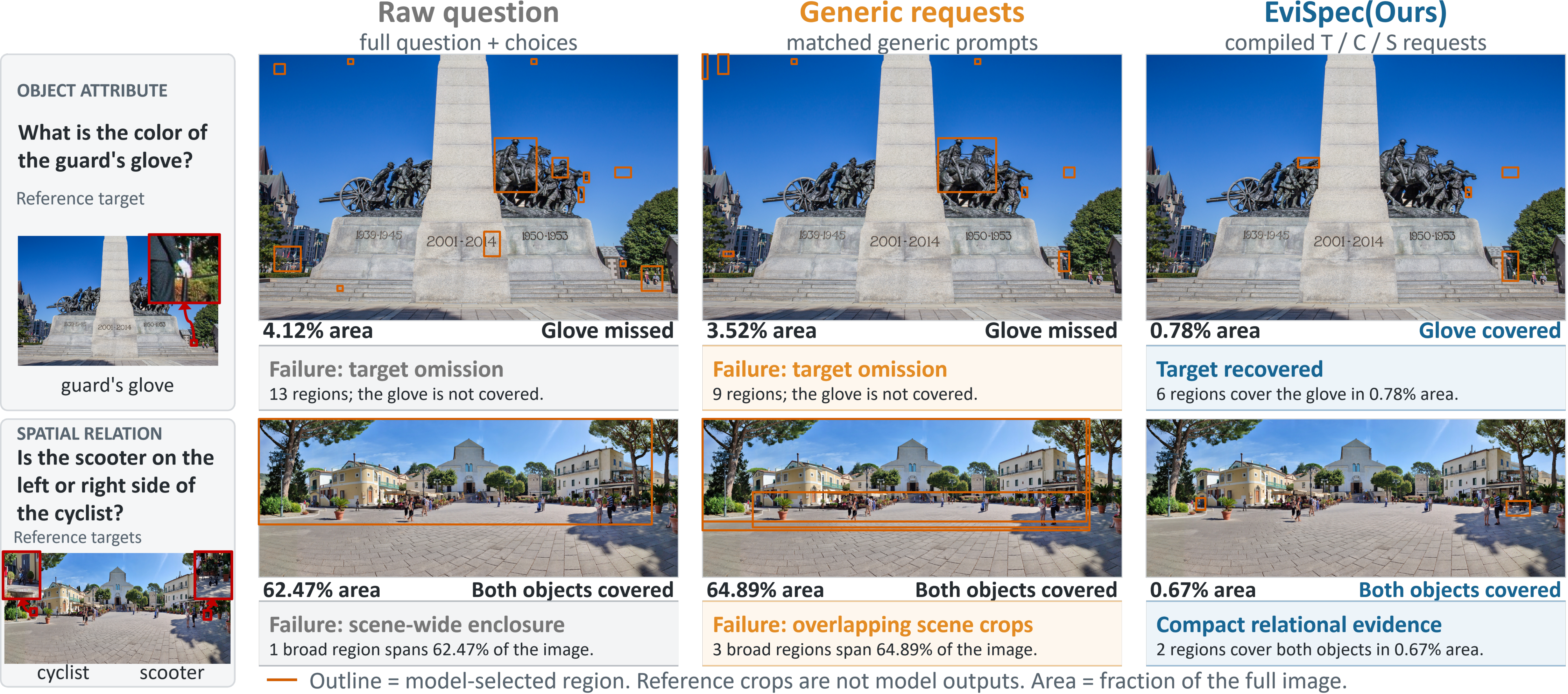}
    \caption{
    Qualitative comparison of evidence localization using raw questions, generic requests.
    }
    \label{fig:qualitative_analysis}
    \vspace{-2mm}
\end{figure*}

\subsection{Quantitative Results}

\textbf{Performance on Benchmarks.} Table~\ref{tab:main_results} compares EviSpec with Base Model, ViCrop, HiDe, DeepEyes~\citep{zheng2025deepeyes}, and DeepScan~\citep{li2026deepscan}. All methods are evaluated under a unified protocol; GPT-4o~\cite{gpt4o} serves as a proprietary reference and is excluded from within-block ranking. EviSpec improves over Base in all 15 model--benchmark combinations, with average relative gains of 10.4\%, 8.8\%, and 12.4\% on V$^*$Bench, HR-Bench-4K, and HR-Bench-8K, respectively. It also outperforms DeepScan across all settings and HiDe across all HR-Bench evaluations.

\textbf{Fine-Grained Gains and Generalization.} The gains span both fine-grained perception and cross-region reasoning. On Qwen2.5-VL-7B, EviSpec improves over Base by 14.8 points on the V$^*$Bench attribute split and 14.3 points on the HR-Bench-8K FCP split. Table~\ref{tab:additional_results} further extends the evaluation to five additional visual tasks, where EviSpec achieves the best aggregate performance across all benchmarks for both models, with particularly strong gains on ZoomBench. Improvements in hallucination evaluation, real-world perception and reasoning, and document understanding further indicate generalization beyond high-resolution visual search.

\subsection{Qualitative Results}
\label{sec:qualitative}
Fig.~\ref{fig:qualitative_analysis} compares localization from raw multiple-choice questions, matched generic requests, and EviSpec. The examples highlight two aspects of evidence selection: recovering small answer-bearing objects and preserving spatial relations. In the upper example, the question asks for the color of a guard's glove in a cluttered monument scene. Raw-question and generic-request localization attend to several salient objects but miss the glove, whereas EviSpec retains it as the target and selects the evidence needed to determine its color. In the lower example, all three conditions cover both the cyclist and scooter, but the raw question and generic requests retain broader scene regions. EviSpec instead selects compact regions around the two entities while preserving their spatial arrangement, retaining the information needed for answering with less surrounding context.

\subsection{Ablation Studies}
\label{sec:ablation}

We evaluate individual specifications and their combinations with Qwen2.5-VL-7B, holding the localization procedure, image budget, and final answering procedure fixed. This ablation assesses the request configuration used for the benchmark evaluation. Table~\ref{tab:ablation} shows that the complete $TCS$ configuration achieves the highest accuracy in all nine reported split and average metrics, reaching 92.1\%, 80.9\%, and 81.6\% on V$^*$Bench, HR-Bench-4K, and HR-Bench-8K, respectively. Among the single-request configurations, $T$ gives the highest average accuracy on all three benchmarks, supporting the use of an explicit answer-bearing target together with its identifying context. Additional perturbation, option-isolation, and fusion analyses are provided in
\appref{sec:app-robustness}.

\section{Conclusion}

This work identifies the evidence-specification gap in high-resolution multimodal reasoning: questions formulated for answering are not effective requests for locating the evidence needed to answer them. We introduce EviSpec, a training-free compiler that separates localization-facing evidence specification from final answering while preserving the original query. Across five frozen MLLMs and three high-resolution benchmarks, EviSpec improves the corresponding base models. More importantly, matched controls show that the gains do not rely on answer-option access or additional region geometry, but on selecting question-relevant visual content. Together, these results show that high-resolution reasoning failures are not solely a problem of visual access, and establish evidence specification as a distinct inference interface between question understanding and visual search.

\bibliography{EviSpec}
\bibliographystyle{iclr2027_conference}

\clearpage
\appendix
\setcounter{secnumdepth}{2}
\setcounter{tocdepth}{2}

\newcommand{\appendixentry}[4]{%
  \noindent
  \begin{tabularx}{\linewidth}{@{}>{\raggedright\arraybackslash}p{0.16\linewidth}X>{\raggedleft\arraybackslash}p{0.13\linewidth}@{}}
    {\large\bfseries\color{gray!60} Appendix~#1} &
    {\large\bfseries #2} &
    {\large\bfseries Page~#4} \\
    & {\normalsize #3} & \\
  \end{tabularx}
  \vspace{0.55em}
}

\thispagestyle{empty}
\begin{center}
  \parbox{0.94\textwidth}{%
    \centering
    {\LARGE\bfseries Appendix: Pay More Attention to Text\\
    in High-Resolution MLLMs\par}%
  }
\end{center}

\vspace{1.6em}

\noindent
\fcolorbox{gray!28}{gray!4}{%
\begin{minipage}{\dimexpr\textwidth-2\fboxsep-2\fboxrule\relax}
\vspace{1.0em}
\hspace{0.7em}{\Large\bfseries Contents}

\vspace{0.9em}
\hspace{0.7em}\rule{\dimexpr\linewidth-1.4em\relax}{0.45pt}
\vspace{0.8em}

\hspace{0.7em}\begin{minipage}{\dimexpr\linewidth-1.4em\relax}
\appendixentry{A}{Relation to Prior Work and Experimental Overview}
{Comparison with related approaches, matched baselines, and an overview of the supplementary evaluations.}
{\pageref{sec:app-closest-work}}

\appendixentry{B}{Matched Question-to-Evidence Controls}
{Prompt-family comparisons, request-interface controls, incremental specification analysis, and component diagnostics.}
{\pageref{sec:app-matched-controls}}

\appendixentry{C}{Matched Selected-Content Intervention}
{Selected-evidence interventions and exact-geometry controls that isolate the contribution of question-relevant visual content.}
{\pageref{sec:app-selected-content}}

\appendixentry{D}{Supplementary Perturbation and Fusion Results}
{Query perturbations, parser audits, option-free controls, compiled-query diversity, fusion studies, and qualitative evidence localization.}
{\pageref{sec:app-robustness}}

\appendixentry{E}{Implementation Details}
{Implementation settings, software and hardware environments, and end-to-end inference-efficiency measurements.}
{\pageref{sec:app-implementation-details}}

\appendixentry{F}{Supplementary Method and Proofs}
{Complete operators, the adaptive-wrapper inference procedure, formal guarantees, and computational-complexity analysis.}
{\pageref{sec:supplement}}
\end{minipage}

\vspace{0.55em}
\end{minipage}%
}

\clearpage

\section{Relation to Prior Work and Experimental Overview}
\label{sec:app-closest-work}

Table~\ref{tab:app-closest-work} compares EviSpec with visual-question
decomposition, question-conditioned cropping, evidence grounding and learned
visual acquisition. EviSpec compiles one multiple-choice question into
complementary localization requests and preserves the complete question for
final answering. Matched comparisons use the same localization and
evidence-view operators across request variants.

Complete-question canonicalization provides a matched baseline for evaluating
role-separated requests and option isolation. The remaining experiments cover
request structure, selected visual content, perturbation consistency,
and the complete inference procedure.

The supplementary evaluation is organized around three evidence blocks.
Prompt-family comparisons evaluate typed request construction. Selected-content
experiments compare the resulting evidence with matched visual alternatives,
while perturbation analyses examine behavior across modified inputs. Implementation
and formal operator details are reported afterward.

\begin{table}[t]
\centering
\caption{
\textbf{Comparison with related approaches.}
``Intermediate object'' denotes the representation produced before final answering.
}
\label{tab:app-closest-work}

{
\small
\renewcommand{\arraystretch}{1.16}
\setlength{\tabcolsep}{3.5pt}
\setlength{\arrayrulewidth}{0.4pt}

\begin{tabularx}{\textwidth}{
    >{\raggedright\arraybackslash}m{0.18\textwidth}
    >{\raggedright\arraybackslash}m{0.22\textwidth}
    >{\centering\arraybackslash}m{0.13\textwidth}
    >{\raggedright\arraybackslash}X
}
\noalign{\hrule height 1pt}

\rowcolor{eviheader}
\textbf{Work}
& \textbf{Intermediate object}
& \textbf{Implementation}
& \textbf{Relation to EviSpec}
\\
\hline

Visual Question Decomposition
\citep{zhang2024vqd}
&
Grounded reasoning subquestions
&
Fine-tuned
&
Decomposes questions into grounded reasoning steps, whereas EviSpec
compiles localization requests while keeping the final reasoning query unchanged.
\\[1mm]

DIEM \citep{jiang2024diem}
&
Matched subquestion--subimage pairs
&
Training-free
&
Pairs subquestions with subimages, while EviSpec produces complementary
localization requests and preserves a single final query.
\\[1mm]

HiDe \citep{liu2025hide}
&
Token-decoupled attention regions and compact layout-preserving evidence
&
Training-free
&
Uses attention-guided evidence regions as a training-free comparator.
EviSpec instead studies how the query is compiled into complementary
requests before a shared localization stage.
\\[1mm]

FOCUS \citep{zhong2025focus}
&
Target-object relevance maps and crops
&
Training-free
&
Improves visual search after identifying target objects, whereas EviSpec
intervenes on the linguistic request supplied to localization.
\\[1mm]

HAVC / LASER
\citep{xie2026havc,zhu2026laser}
&
Head- or layer-selected attention crops
&
Training-free
&
Improve which internal spatial signal is used; EviSpec instead changes
the query representation evaluated through a shared spatial operator.
\\[1mm]

AdaptVision \citep{lin2026adaptvision}
&
Learned crop-tool actions
&
Reinforcement learning
&
Learns an acquisition policy and token allocation, whereas EviSpec uses
a training-free request compiler.
\\[1mm]

Visual Grounding for Object Questions
\citep{everaert2026vgoq}
&
Answer-supporting evidence regions
&
Trained grounding model
&
Studies a broader grounding task, while EviSpec focuses on
question-to-evidence compilation for high-resolution visual reasoning.
\\

\noalign{\hrule height 1pt}
\end{tabularx}
}
\end{table}

\section{Matched Question-to-Evidence Controls}
\label{sec:app-matched-controls}

\subsection{Request Interface and Matched Prompt-Family Controls}
\label{sec:app-prompt-controls}
Three controlled evaluations distinguish request structure from downstream
visual processing. All use the 191 V$^*$Bench questions, Qwen2.5-VL-7B,
layer~15, bfloat16, FlashAttention-2, a fixed $448\times448$ evidence-view
resolution, the original image and unchanged final question. Request generation
permits at most 1,024 new tokens; answer generation uses the same greedy,
valid-letter contract in every condition.

\subsubsection{Prompt templates and request mapping}

The T/C/S templates and matched generic templates are held constant throughout
the matched comparison. Table~\ref{tab:app-prompt-map} gives all six request
instances. $C$ and $S$ share one instruction literal and differ only in whether
the substituted input is the complete serialized question $Q$ or the
option-free prefix $q_0$.

\begin{table}[ht]
\centering
\small
\caption{Request-by-request mapping in the primary matched prompt-family
control. The two families have identical input scopes and output tags.}
\label{tab:app-prompt-map}
\begin{tabular}{llll}
\toprule
Family & Request & Input & Required tags \\
\midrule
T/C/S & $T$ & complete $Q$ & \texttt{MODE}, \texttt{FINAL\_OUTPUT} \\
T/C/S & $C$ & complete $Q$ & \texttt{FINAL\_OUTPUT} \\
T/C/S & $S$ & option-free $q_0$ & \texttt{FINAL\_OUTPUT} \\
Generic & $G_1$ & complete $Q$ & \texttt{MODE}, \texttt{FINAL\_OUTPUT} \\
Generic & $G_2$ & complete $Q$ & \texttt{FINAL\_OUTPUT} \\
Generic & $G_3$ & option-free $q_0$ & \texttt{FINAL\_OUTPUT} \\
\bottomrule
\end{tabular}
\end{table}

The operational lexical content of the five unique instruction literals is
reproduced below, with Markdown emphasis removed and line wrapping normalized
for typesetting; \texttt{\{INPUT\}} marks the runtime substitution. The
complete prompt text and request pairing are therefore specified directly in
this appendix.

\paragraph{Role-separated $T$ literal.}
\begin{quote}\footnotesize
You are a visual grounding query parser. Do not answer the multiple-choice
question and do not copy answer options.

First decide whether the question is DIRECT or RELATION.

DIRECT asks for an attribute, text, count, identity, or existence. Output one
answer-bearing target phrase. Preserve identifying context inside the same
phrase. Example: ``What color is the backpack carried by the man wearing a
yellow shirt?'' $\rightarrow$ backpack carried by man wearing yellow shirt.

RELATION asks to compare, order, match, or locate one visual object relative to
another. Output both the subject and reference as two independently groundable
noun phrases, in question order. Never omit the subject. Example: ``Where is
the person in relation to the recycle bin?'' $\rightarrow$ person, recycle bin.
Example: ``What is the position of the red car compared to the black car?''
$\rightarrow$ red car, black car.

Return exactly two tagged fields and no other text:
\texttt{<MODE>direct</MODE><FINAL\_OUTPUT>one target phrase</FINAL\_OUTPUT>}
or
\texttt{<MODE>relation</MODE><FINAL\_OUTPUT>subject, reference</FINAL\_OUTPUT>}.

Use lowercase. A comma followed by a space is reserved for separating RELATION
targets. Question: \texttt{\{INPUT\}}.
\end{quote}

\paragraph{Canonical-form $C/S$ literal.}
\begin{quote}\footnotesize
You are a highly precise language analysis engine. Your sole function is to
extract entities (e.g., objects, people) from a user's question, and deconstruct
them into a canonical, attribute-based format by strictly following a set of
rules and a thinking process.

\textbf{Thinking Process.} Before generating the final output, you must
internally follow these steps in order:

1. Identify Core Entities: read the entire question and identify all key noun
phrases, for example, ``the green surfboard'' and ``the purple umbrella''.

2. Deconstruct Attributes for Each Entity Individually: before considering the
relationship between entities, look at each entity in isolation and apply
Rules 2, 3 and 4 to fully deconstruct its attributes. For instance, first
process ``the green surfboard'' using Rule 2 to obtain
\texttt{surfboard with green color}, then process ``the purple umbrella'' to
obtain \texttt{umbrella with purple color}.

3. Handle Relationships Between Entities: after all entities have been
individually deconstructed, check for spatial or logical relationships between
them (Rule 5). If a relationship exists, list the already deconstructed
entities as separate items.

4. Assemble and Normalize: gather all canonical entity strings, convert all
text to lowercase, and join all entities into a single line separated by a
comma and a space.

5. Final Formatting: enclose the resulting single-line string within the
\texttt{<FINAL\_OUTPUT>} and \texttt{</FINAL\_OUTPUT>} tags.

\textbf{Extraction Rules.}
Rule 1 (Simple Entities): if a noun is not described by modifiers, extract the
noun itself; for example, ``the scooter'' becomes \texttt{scooter}.
Rule 2 (Adjective Attribute Deconstruction): if an entity is modified by one or
more adjectives, use
\texttt{noun with [property] [type]}, chaining multiple properties. Colors map
to \texttt{color}, sizes to \texttt{size}, and materials to
\texttt{material}; for example, ``the large blue truck'' becomes
\texttt{truck with large size with blue color}.
Rule 3 (Possessive Inversion): convert possessive forms such as
\texttt{X's Y} to \texttt{Y of X}; for example, ``the woman's handbag''
becomes \texttt{handbag of woman}.
Rule 4 (Attributive Prepositional Phrases): if a prepositional phrase describes
a component of an entity, preserve the structure and recursively apply the
rules to the entity within the phrase; for example, ``the man in the green
shirt'' becomes \texttt{man in a shirt with green color}.
Rule 5 (Relational Prepositional Phrases): if a prepositional phrase describes
a relationship between two separate entities, extract the entities as separate
items only after each has been fully deconstructed. Do not include relational
words. Thus, ``the dog on the left side of the scooter'' becomes
\texttt{dog, scooter}, and ``Is the green surfboard on the left side of the
purple umbrella?'' becomes
\texttt{surfboard with green color, umbrella with purple color}.
Rule 6 (Compound Nouns): treat recognized compound nouns as one entity; for
example, ``the traffic light'' becomes \texttt{traffic light}.

\textbf{Output Format Rules.} The final and only output content must be enclosed
within \texttt{<FINAL\_OUTPUT>} and \texttt{</FINAL\_OUTPUT>}. It must be one
continuous lowercase line. Separate multiple entities by a comma followed by a
space. Exclude articles, question words and purely relational words.

Now, following all the rules above, extract the entities from the question
below: \texttt{\{INPUT\}}.
\end{quote}

\paragraph{Matched generic literals.}
\begin{quote}\footnotesize
$G_1$: You are a visual search query rewriter. Do not answer the
multiple-choice question. First decide whether the input is DIRECT (attribute,
text, count, identity, or existence) or RELATION (comparison, order, match, or
location of one object relative to another). Then write a concise visual search
description that preserves every visual object, attribute, and relation without
assigning target, context, or spatial roles or adding facts. For RELATION
include both compared objects in question order. Return exactly
\texttt{<MODE>direct</MODE><FINAL\_OUTPUT>one search description</FINAL\_OUTPUT>}
or
\texttt{<MODE>relation</MODE><FINAL\_OUTPUT>one search description containing
both objects</FINAL\_OUTPUT>}. Input: \texttt{\{INPUT\}}.

$G_2$: Paraphrase the input below into one concise visual search description.
Preserve every visual object, attribute, and relation; do not add facts or
assign semantic roles. Return only
\texttt{<FINAL\_OUTPUT>one search description</FINAL\_OUTPUT>}.
Input: \texttt{\{INPUT\}}.

$G_3$: Convert the input below into one concise visual search description.
Preserve every visual object, attribute, and relation; do not add facts or
assign semantic roles. Return only
\texttt{<FINAL\_OUTPUT>one search description</FINAL\_OUTPUT>}.
Input: \texttt{\{INPUT\}}.
\end{quote}

Table~\ref{tab:app-prompt-examples} illustrates the requests produced by the
two prompt families. Both complete-input families can reproduce option text,
and each matched variant explicitly specifies the text available to the request
generator.

\begin{table}[ht]
\centering
\small
\setlength{\tabcolsep}{3pt}
\caption{Example requests generated by the structured and matched generic
prompt families. Entries are ordered $T/C/S$ or $G_1/G_2/G_3$.}
\label{tab:app-prompt-examples}
\begin{tabular}{p{0.06\textwidth}p{0.42\textwidth}p{0.46\textwidth}}
\toprule
ID & T/C/S outputs & Matched generic outputs \\
\midrule
0 & \texttt{glove}; \texttt{glove with rubber material};
\texttt{glove} &
\texttt{glove material}; \texttt{find glove material: rubber, cotton, kevlar,
leather.}; \texttt{visual search for glove material} \\
1 & \texttt{dustpan}; \texttt{dustpan with purple color};
\texttt{dustpan with color} &
\texttt{the color of the dustpan}; \texttt{find a dustpan with a purple
color.}; \texttt{visual search for dustpan with specific color} \\
\bottomrule
\end{tabular}
\end{table}

The experiments below use frozen Qwen2.5-VL-7B analysis archives with the
same image processor, answer prompt and greedy decoding setup. The main table
reports cross-model benchmark results, while these matched executions examine
request construction and visual evidence. Accuracies are rounded to one decimal
place.

``Direct'' denotes the Base answer pipeline, ``T/C/S'' denotes the complete
EviSpec request construction, and ``Selected evidence'' denotes the visual
content supplied by its matched intervention. The main-table EviSpec accuracy is
92.1\%, while the matched request-construction and selected-content executions
reach 90.6\% and 91.6\%, respectively.

\subsection{Request-interface comparisons}
\label{sec:app-request-interface}

The first comparison holds the visual operator and answer procedure fixed while
changing only the requests supplied to localization. Direct uses the complete
question without an evidence request. The option-free stem removes the answer
options, and canonical $C$ rewrites the complete question into a structured
entity description. The T/C/S compiler emits a role-separated target request,
a canonical entity request and an option-isolated request. A union control packs
these regions into one canvas, whereas the standard T/C/S pipeline keeps the
views separate.

\begin{table}[t]
\centering
\small
\caption{Matched request-interface comparison on 191 V$^*$Bench questions. Accuracy values are percentages rounded to one decimal place; bold marks the higher value in each row.}
\label{tab:app-interface-attribution}
\begin{tabular}{lrr}
\toprule
Comparator & Comparator accuracy & T/C/S accuracy \\
\midrule
Raw option-free stem & 83.2 & \textbf{88.5} \\
Canonical $C$ & 89.0 & \textbf{90.1} \\
\bottomrule
\end{tabular}
\end{table}

The one-canvas T/C/S union reaches 88.5\% against the raw option-free stem and
90.1\% against canonical $C$. The separate-view execution reaches 90.6\% on the
same 191 V$^*$Bench questions. The one-canvas comparison evaluates request
construction, while the EviSpec pipeline presents separate evidence views to
the final answer stage.

\begin{table}[t]
\centering
\small
\caption{Equal-search-budget comparison. Both conditions attempt three generated requests before duplicate removal and evidence assembly. Bold marks higher accuracy or exclusive successes, and fewer distinct requests.}
\label{tab:app-search-budget}
\begin{tabular}{lrr}
\toprule
Quantity & Generic3 & T/C/S \\
\midrule
Accuracy & 79.1 & \textbf{88.5} \\
Exclusive successes & 4 & \textbf{22} \\
Average distinct non-empty requests & 2.6 & \textbf{1.9} \\
\bottomrule
\end{tabular}
\end{table}

The structured compiler uses fewer distinct requests and reaches higher accuracy
under the same request budget. Both conditions receive three generation
attempts; T/C/S assigns target, canonical and option-isolated roles to them.

\begin{table}[t]
\centering
\small
\caption{Request-by-request prompt-family comparison on V$^*$Bench. Bold marks higher accuracy or exclusive successes, fewer requests, and smaller selected area.}
\label{tab:app-role-matched}
\begin{tabular}{lrrr}
\toprule
Quantity & Direct & Matched generic & T/C/S \\
\midrule
Accuracy & 78.5 & 77.0 & \textbf{84.3} \\
Exclusive successes & -- & 7 & \textbf{21} \\
Average distinct non-empty requests & -- & 2.9 & \textbf{1.9} \\
Three distinct requests & -- & 176/191 & 27/191 \\
Average selected area & -- & 69.6 & \textbf{8.0} \\
\bottomrule
\end{tabular}
\end{table}

The matched generic family retains more distinct requests and a larger selected
area, whereas the structured family produces higher accuracy with fewer, more
compact requests. Model capacity and the downstream visual operator are unchanged
between the two prompt families.

\begin{figure}[t]
\centering
\includegraphics[width=0.96\textwidth]{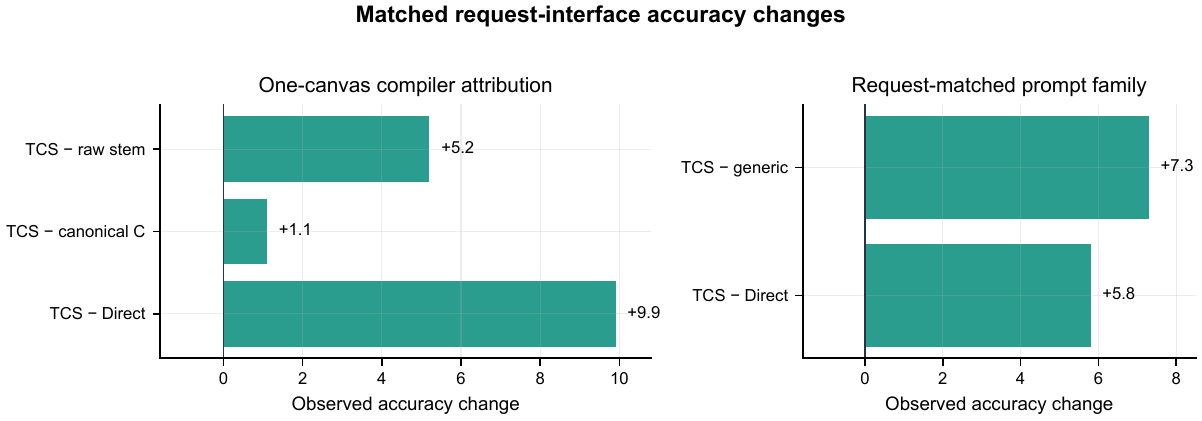}
\caption{Matched request-interface accuracy changes. The left panel reports the one-canvas compiler-attribution contrasts; the right panel reports the request-matched prompt-family contrasts.}
\label{fig:app-interface-attribution}
\end{figure}

\subsection{Incremental specification and component diagnostics}
\label{sec:app-incremental}

The branch diagnostic derives each accuracy change from the controlled request
ablation in the main paper. It records the effect of adding the canonical or
option-isolated request to an existing request set.

\begin{table}[t]
\centering
\small
\caption{Observed incremental specification changes in accuracy. Within each benchmark, bold marks the largest positive change.}
\label{tab:app-ablation-paired}
\begin{tabular}{llr}
\toprule
Benchmark & Contrast & Accuracy change \\
\midrule
V$^*$Bench & $TC-T$ & $+0.5$ \\
 & $TS-T$ & $+1.0$ \\
 & $TCS-TC$ & $\mathbf{+3.1}$ \\
 & $TCS-TS$ & $+2.6$ \\
\midrule
HR-Bench-4K & $TC-T$ & $+0.7$ \\
 & $TS-T$ & $+0.7$ \\
 & $TCS-TC$ & $\mathbf{+1.9}$ \\
 & $TCS-TS$ & $\mathbf{+1.9}$ \\
\midrule
HR-Bench-8K & $TC-T$ & $-0.5$ \\
 & $TS-T$ & $-0.2$ \\
 & $TCS-TC$ & $\mathbf{+3.5}$ \\
 & $TCS-TS$ & $+3.2$ \\
\bottomrule
\end{tabular}
\end{table}

Across all three benchmarks, the largest increase occurs when the final missing
request is added to form TCS. This pattern complements the main ablation by
showing that the three request roles provide their strongest result jointly.

\begin{figure}[t]
\centering
\includegraphics[width=0.96\textwidth]{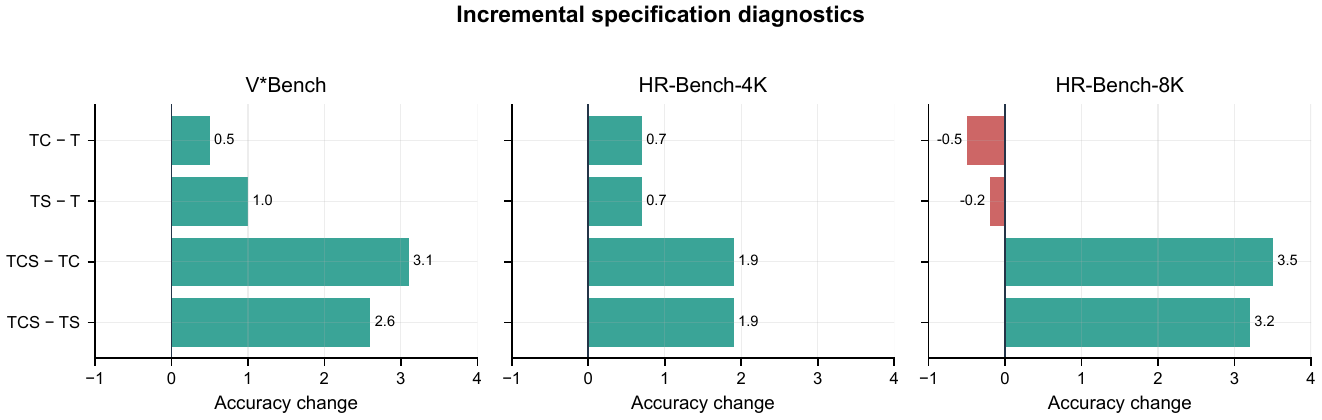}
\caption{Observed incremental specification changes across the three benchmarks. Zero marks unchanged accuracy.}
\label{fig:app-ablation-paired}
\end{figure}

\begin{table}[t]
\centering
\small
\caption{Matched component diagnostic on 191 V$^*$Bench questions. All entries are percentages; localization columns report region measurements. Bold marks the highest value in each metric column.}
\label{tab:app-factorial}
\begin{tabular}{llrrrr}
\toprule
Attention map & Region formation & Accuracy & mIoU & mIoG & Cov.@0.5 \\
\midrule
Legacy & Legacy & 89.5 & 12.9 & \textbf{93.5} & \textbf{94.2} \\
Legacy & Adaptive & 86.4 & 1.6 & 92.7 & 92.7 \\
Calibrated & Legacy & 85.3 & \textbf{28.8} & 50.3 & 48.2 \\
Calibrated & Adaptive & \textbf{93.2} & 17.8 & 92.6 & 93.7 \\
\bottomrule
\end{tabular}
\end{table}

The component grid separates attention-map processing from region formation.
Calibrated maps with adaptive regions achieve the highest accuracy at 93.2 and
raise mIoU from 12.9 to 17.8 relative to the legacy pairing. The legacy pairing
retains the highest mIoG and coverage. EviSpec uses the calibrated-map and
adaptive-region configuration in the request-construction comparisons.

\begin{figure}[t]
\centering
\includegraphics[width=0.98\textwidth]{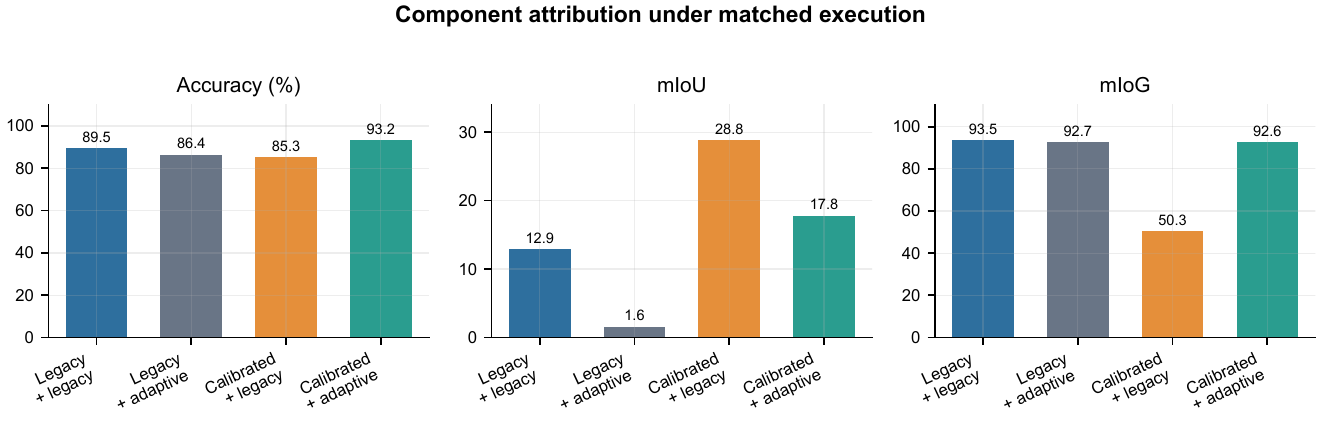}
\caption{Component attribution under matched execution. The panels show accuracy, mIoU and mIoG for the four attention-map and region-formation pairings.}
\label{fig:app-factorial}
\end{figure}

\section{Matched Selected-Content Intervention}
\label{sec:app-selected-content}

This intervention keeps the frozen model, answer prompt, layer and question set
fixed while changing the visual content supplied to the model. Blank and resized
views control the image budget, selected-only input measures evidence
sufficiency, selected-region masking removes the selected appearance, and random
controls preserve the number and geometry of selected boxes.

\begin{center}
\refstepcounter{table}\label{tab:app-evidence-intervention}
\small
\textbf{Table \thetable: Matched V$^*$Bench evidence-intervention results.} Values are observed accuracy percentages; bold marks the highest value.\\[2pt]
\begin{tabular}{lr}
\toprule
Condition & Accuracy (\%) \\
\midrule
Direct & 78.5 \\
Blank evidence views & 78.5 \\
Global resized views & 79.1 \\
Random evidence & 80.5 \\
Selected evidence & \textbf{91.6} \\
Selected evidence only & \textbf{91.6} \\
Random mask & 79.8 \\
Selected-region mask & 44.0 \\
\bottomrule
\end{tabular}
\end{center}

Selected evidence produces the highest observed accuracy among the content
conditions, while masking the selected region removes the appearance cues needed
for reliable answering. The selected-only condition matches selected evidence in
this execution, so the compact evidence view alone carries the visual content
used for answering.

\begin{figure}[t]
\centering
\includegraphics[width=0.98\textwidth]{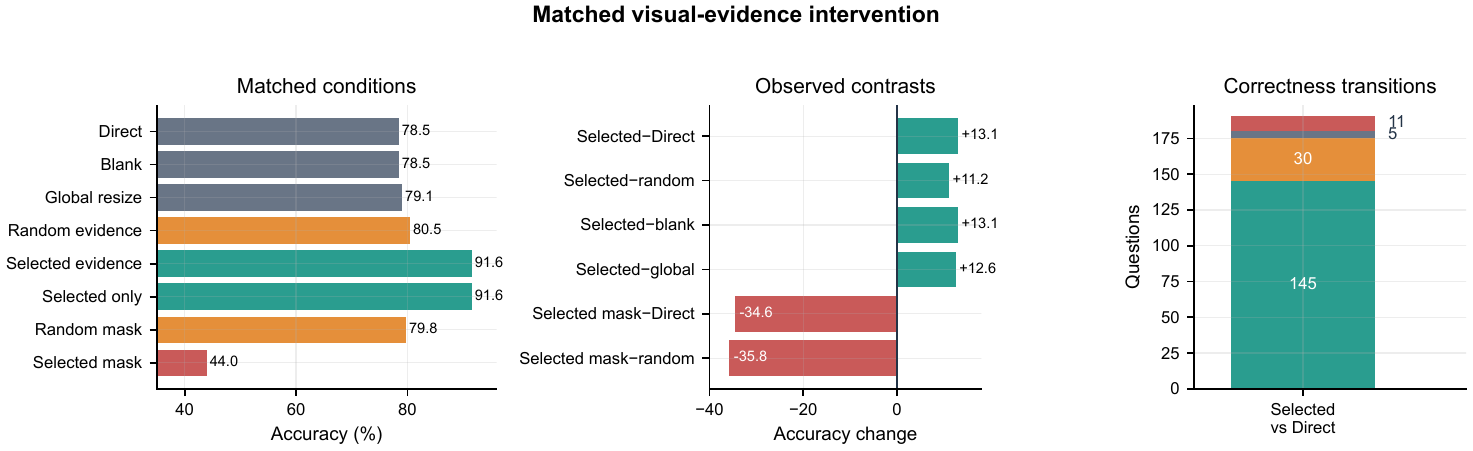}
\caption{Matched visual-evidence intervention. The panels show condition accuracies, accuracy differences between matched conditions and correctness transitions.}
\label{fig:app-evidence-intervention}
\end{figure}

\subsection{Exact-geometry rigid-translation confirmation}

The geometry-matched control translates the complete selected box configuration
with one shared integer-pixel offset. It preserves box count, pixel dimensions,
overlap, union area and packed-view dimensions while changing the source content.

\begin{center}
\refstepcounter{table}\label{tab:app-exact-geometry}
\small
\textbf{Table \thetable: Exact-geometry V$^*$Bench conditions.} Values are observed accuracy percentages; bold marks the highest value.\\[2pt]
\begin{tabular}{lr}
\toprule
Condition & Accuracy (\%) \\
\midrule
Direct & 78.5 \\
Selected evidence & \textbf{91.6} \\
Exact-geometry random evidence & 79.8 \\
Selected-region mask & 43.5 \\
Exact-geometry random mask & 77.1 \\
\bottomrule
\end{tabular}
\end{center}

All translated configurations preserve the assembled-view geometry by
construction. The selected-evidence condition remains higher than the
geometry-matched random evidence condition, while the selected-region mask
remains lower than its random-mask counterpart. This isolates source content as
the changing factor in the matched control.

\begin{figure}[t]
\centering
\includegraphics[width=0.98\textwidth]{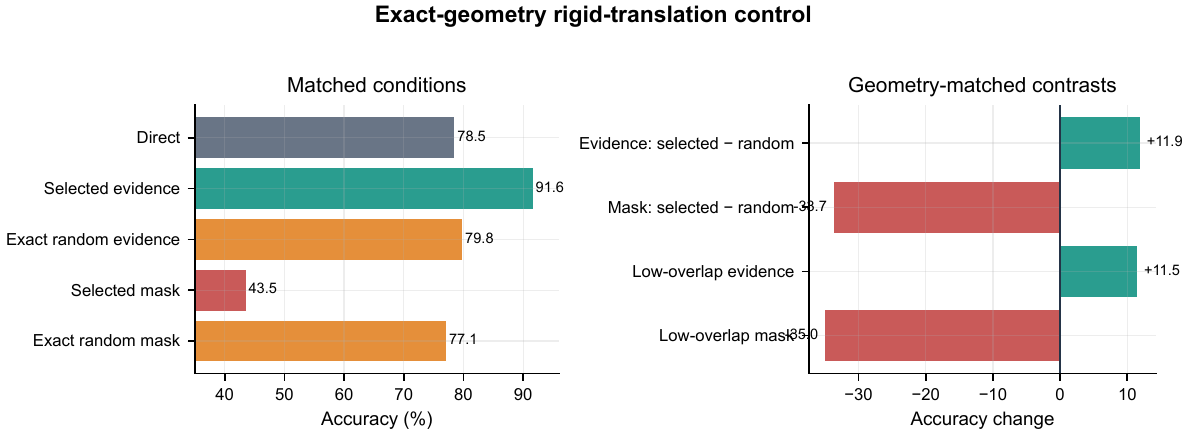}
\caption{Exact-geometry rigid-translation control. The panels report condition accuracies and accuracy differences for the full set and the low-overlap subset.}
\label{fig:app-exact-geometry}
\end{figure}

\section{Supplementary Perturbation and Fusion Results}
\label{sec:app-robustness}

\subsection{Query perturbations and parser audits}

The perturbation suite applies option permutations, distractor replacements and
target rewrites to the same V$^*$Bench questions. Higher values are preferable
for evidence IoU, worst-case accuracy and all-correct accuracy; lower values are
preferable for answer changes.

\begin{table*}[t]
\centering
\small
\caption{Question-perturbation results on V$^*$Bench. All entries are percentages; bold marks the better value within each selector pair.}
\label{tab:query_robustness}
\begin{tabular}{llrrrr}
\toprule
Selector & Query strategy & Evidence IoU & Answer change & Worst accuracy & All correct \\
\midrule
Fixed-threshold & Single-$C$ & 82.2 & 1.7 & 89.7 & 87.4 \\
Fixed-threshold & $TCS$ & \textbf{84.0} & \textbf{0.9} & \textbf{91.6} & \textbf{90.1} \\
Adaptive & Single-$C$ & 81.1 & 2.2 & 86.9 & 83.8 \\
Adaptive & $TCS$ & \textbf{81.9} & \textbf{1.8} & \textbf{89.5} & \textbf{86.9} \\
\bottomrule
\end{tabular}
\end{table*}

Under both selectors, TCS gives higher evidence IoU, worst-case accuracy and
all-correct accuracy, together with fewer answer changes than Single-$C$.

\begin{table}[t]
\centering
\small
\caption{Parser diagnostics over 1,348 original and perturbed records. Entries are percentages; bold marks the lowest value in each column.}
\label{tab:app-parser-diagnostics}
\begin{tabular}{lrr}
\toprule
Parser input & Empty output (\%) & Complete option overlap (\%) \\
\midrule
$T$-full & 4.4 & 2.5 \\
$T$-stem & \textbf{0.3} & \textbf{1.8} \\
$C$ & 1.3 & 43.5 \\
$S$ & 1.3 & 6.2 \\
\bottomrule
\end{tabular}
\end{table}

The target-stem branch has the lowest empty-output rate and the lowest complete
option-overlap rate. Canonical $C$ retains the complete question and therefore
shows the largest option overlap.

\begin{figure}[t]
\centering
\includegraphics[width=0.96\textwidth]{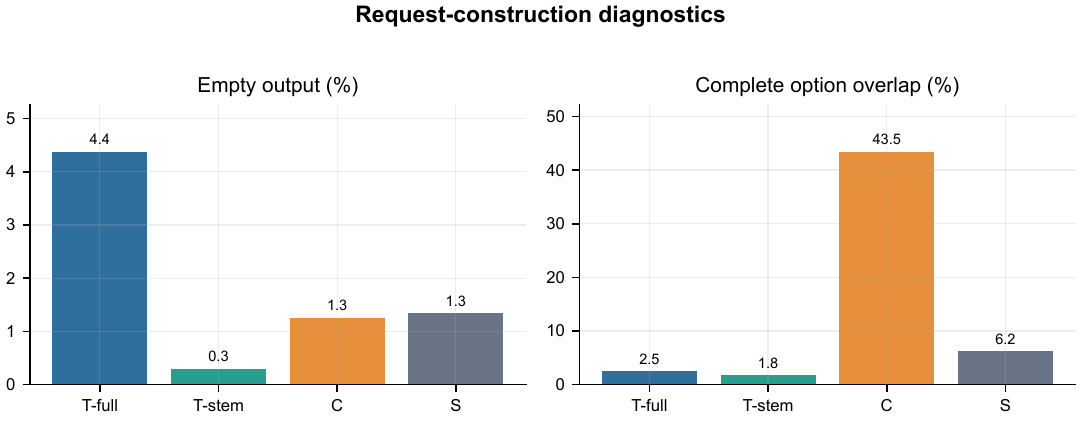}
\caption{Request-construction diagnostics. The panels report empty-output rates and complete answer-option overlap for the four parser inputs.}
\label{fig:app-option-overlap}
\end{figure}

\begin{figure}[t]
\centering
\includegraphics[width=0.96\textwidth]{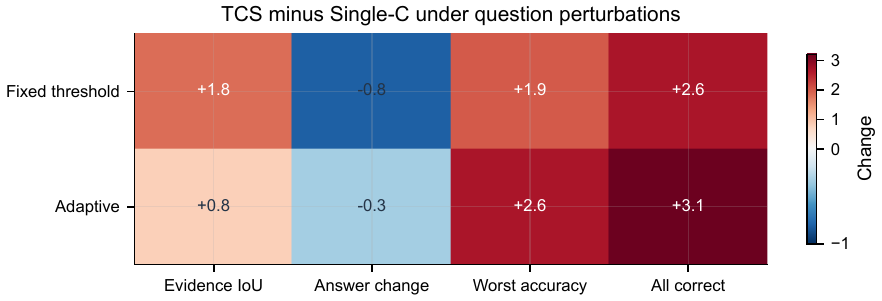}
\caption{Changes from Single-$C$ to TCS under question perturbations. Positive values indicate higher evidence stability or accuracy; a negative answer-change value indicates fewer changed answers.}
\label{fig:app-robustness}
\end{figure}

\subsection{Strict option-free compiler intervention}

The option-free intervention removes answer options before request construction
while keeping the model, layer, image budget and answer prompt fixed.

\begin{table}[t]
\centering
\small
\caption{Matched strict option-free control on V$^*$Bench. Accuracy values are percentages rounded to one decimal place; bold marks the best value in each column.}
\label{tab:app-option-free}
\begin{tabular}{lrrrr}
\toprule
Method & Correct & Overall accuracy & Attribute & Spatial \\
\midrule
Direct & 150/191 & 78.5 & 80.0 & 76.3 \\
Single-$C$ & 168/191 & 88.0 & 91.3 & 82.9 \\
Single-$T$-stem & 169/191 & 88.5 & \textbf{93.0} & 81.6 \\
Single-$S$ & 168/191 & 88.0 & 91.3 & 82.9 \\
Strict option-free $TS$ & \textbf{171/191} & \textbf{89.5} & \textbf{93.0} & \textbf{84.2} \\
Full-input $TCS$ & \textbf{171/191} & \textbf{89.5} & \textbf{93.0} & \textbf{84.2} \\
\bottomrule
\end{tabular}
\end{table}

The strict option-free compiler reaches the same aggregate accuracy as the
full-input TCS condition in this matched execution. The correctness transition
count is 170 correct-to-correct, 19 incorrect-to-incorrect, and one exchange in
each direction. On HR-Bench-8K, adding $C$ to $TS$ raises accuracy by 3.2 in
Table~\ref{tab:app-ablation-paired}; the contribution of $C$ therefore varies
with the benchmark.

\begin{figure}[t]
\centering
\includegraphics[width=0.96\textwidth]{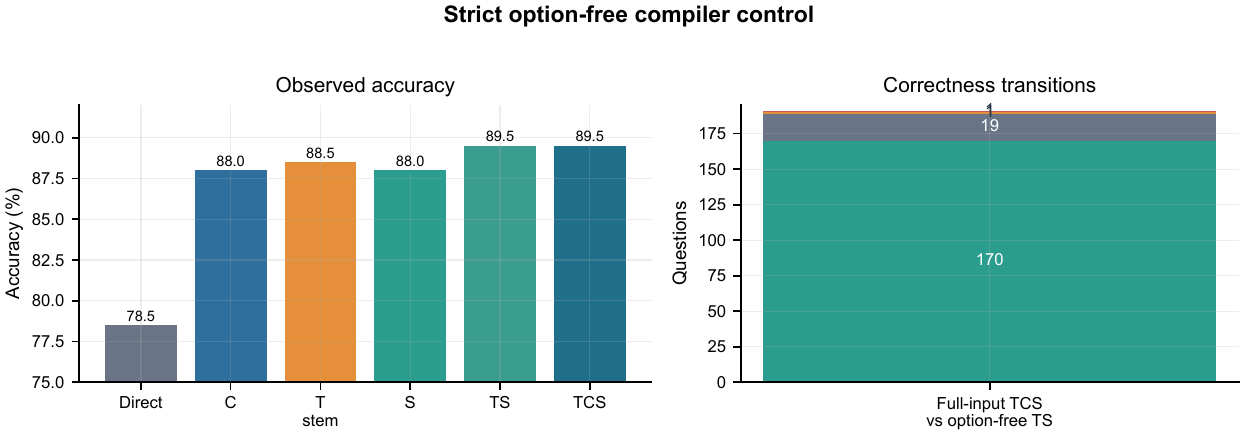}
\caption{Strict option-free compiler control. The left panel reports observed accuracy and the right panel shows correctness transitions between full-input TCS and option-free TS.}
\label{fig:option-free}
\end{figure}

\subsection{Compiled-query diversity and fusion}

\begin{table}[t]
\centering
\small
\caption{Number of distinct, non-empty compiled queries on 191 unperturbed V$^*$Bench questions. The final three columns are percentages within each row.}
\label{tab:app-query-diversity}
\begin{tabular}{lrrrr}
\toprule
Question group & $N$ & One query & Two queries & Three queries \\
\midrule
All & 191 & 20.9 & 64.9 & 14.1 \\
Attributes & 115 & 1.7 & 76.5 & 21.7 \\
Spatial & 76 & 50.0 & 47.4 & 2.6 \\
\bottomrule
\end{tabular}
\end{table}

The query count depends on question type. Attribute questions usually retain two
or three complementary requests, whereas spatial questions more often collapse
to one or two requests.

\begin{table}[t]
\centering
\small
\renewcommand{\arraystretch}{1.0}
\setlength{\tabcolsep}{1.35mm}
\caption{Observed fusion comparison on 191 unperturbed V$^*$Bench questions. All entries are percentages; bold marks the best value in each metric within each selector block.}
\label{tab:app-fusion-complete}
\begin{tabular}{llrrrrrr}
\toprule
Selector & Fusion strategy & Acc. & Attr & Spatial & mIoU & mIoG & Cov.@0.5 \\
\midrule
Fixed-threshold & Single-$T$ & 90.6 & 93.9 & 85.5 & 11.1 & 83.6 & 84.1 \\
Fixed-threshold & Single-$C$ & 92.2 & 95.7 & 86.8 & 11.2 & 85.4 & 86.1 \\
Fixed-threshold & Single-$S$ & 92.2 & 95.7 & 86.8 & 11.2 & 84.6 & 85.3 \\
Fixed-threshold & Naive union & 89.5 & 93.0 & 84.2 & 10.7 & 87.0 & 87.8 \\
Fixed-threshold & Attention average & 91.1 & 95.7 & 84.2 & 12.0 & 86.1 & 87.4 \\
Fixed-threshold & Confidence selection & 93.2 & 97.4 & 85.5 & 11.4 & 85.7 & 86.5 \\
Fixed-threshold & $TCS$ concat & \textbf{93.7} & \textbf{98.3} & \textbf{88.2} & \textbf{12.5} & \textbf{91.5} & \textbf{92.2} \\
\midrule
Adaptive & Single-$T$ & 88.0 & 92.2 & 81.6 & 14.4 & 80.5 & 82.9 \\
Adaptive & Single-$C$ & 88.0 & 91.3 & 82.9 & 10.9 & 84.7 & 85.7 \\
Adaptive & Single-$S$ & 88.0 & 91.3 & 82.9 & 14.0 & 82.6 & 84.5 \\
Adaptive & Naive union & 88.0 & 91.3 & 82.9 & 10.7 & 86.4 & 87.4 \\
Adaptive & Attention average & 85.3 & 91.3 & 76.3 & 7.0 & 65.4 & 66.5 \\
Adaptive & Confidence selection & \textbf{89.5} & 93.0 & 81.6 & 12.8 & 84.7 & 86.5 \\
Adaptive & $TCS$ concat & \textbf{89.5} & \textbf{94.8} & \textbf{84.2} & \textbf{15.8} & \textbf{90.9} & \textbf{91.8} \\
\bottomrule
\end{tabular}
\end{table}

For the fixed-threshold selector, TCS concatenation achieves the highest value
in every reported column. For the adaptive selector, it ties the highest
accuracy and achieves the highest attribute, spatial, mIoU, mIoG and coverage
values.

\begin{figure}[t]
\centering
\includegraphics[width=0.96\textwidth]{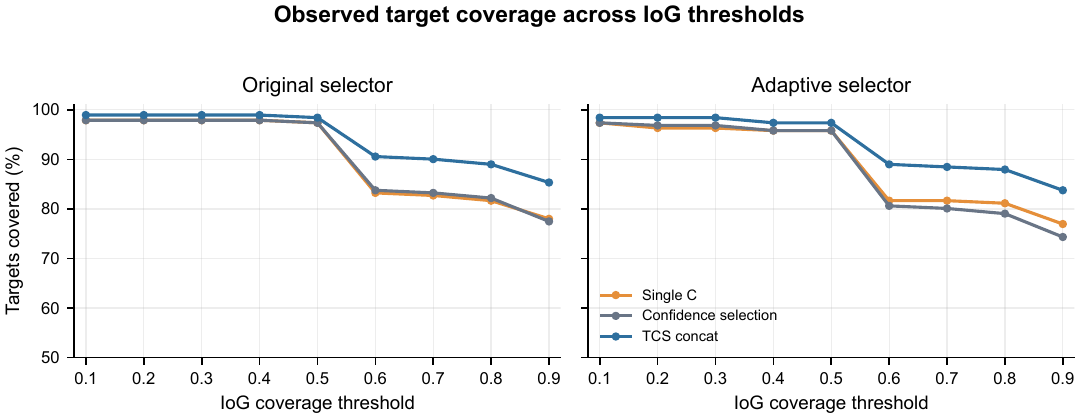}
\caption{Observed target coverage across matched IoG thresholds.}
\label{fig:app-iog-thresholds}
\end{figure}

\subsection{Qualitative evidence localization}

\begin{figure}[t]
\centering
\includegraphics[width=0.99\textwidth]{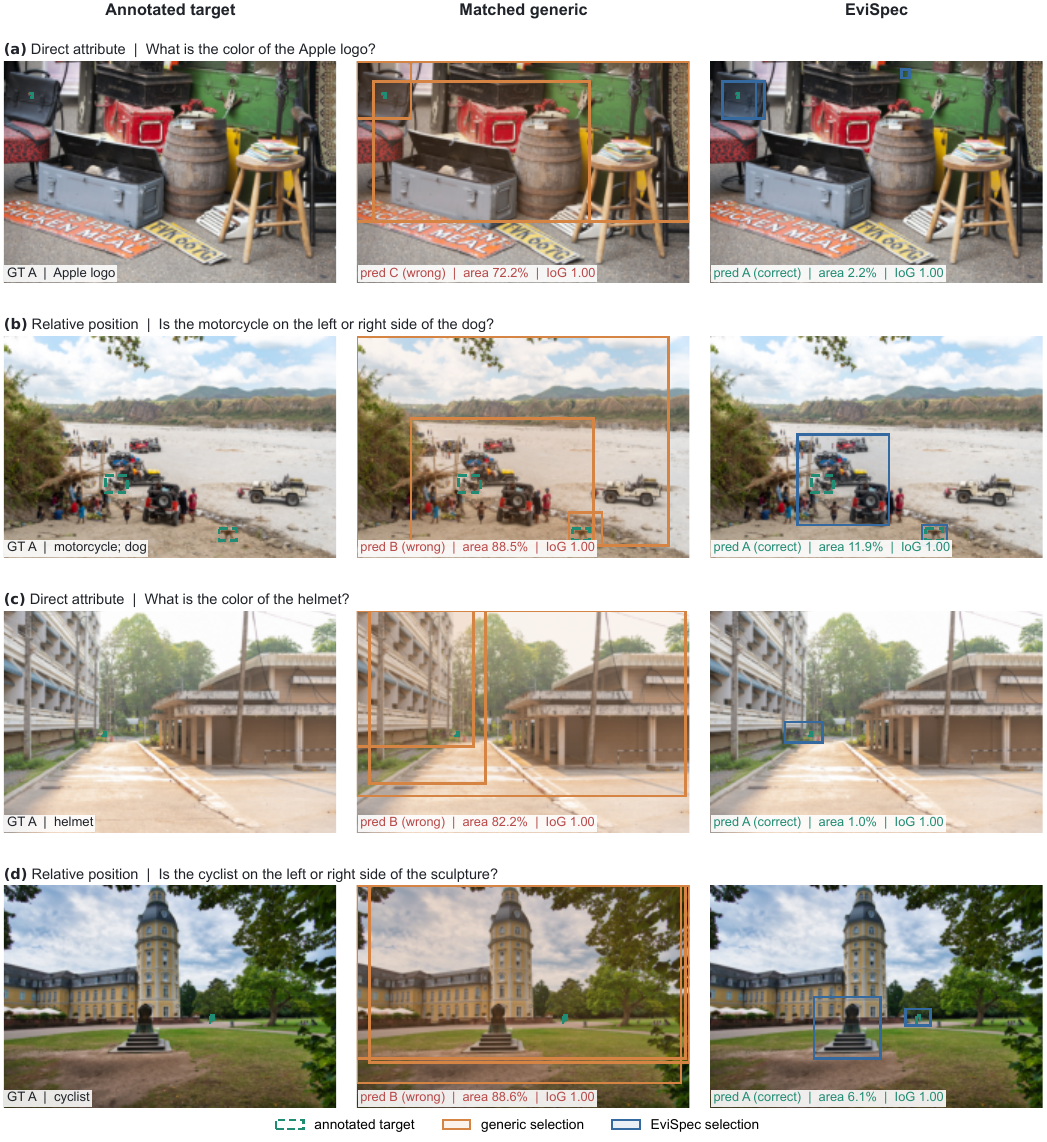}
\caption{Qualitative comparison of evidence localization on V$^*$Bench. Each row shows the original image, the selected regions and the assembled evidence view for representative attribute and relation questions.}
\label{fig:app-qualitative-examples}
\end{figure}

\section{Implementation Details}
\label{sec:app-implementation-details}

\subsection{Implementation settings}

All Qwen2.5-VL diagnostic controls in this appendix use
\texttt{USE\_ALLATT=False}, bfloat16, FlashAttention-2, greedy decoding and the
same $16{,}384\times28^2$ maximum processor budget. The software environment
uses Python 3.11.13, PyTorch 2.8.0+cu128 and Transformers 4.57.1, and experiments
are run on NVIDIA A800 GPUs. The same compiler, attention-processing,
region-formation and evidence-conditioned answering procedures are used across
datasets unless otherwise stated.

\subsection{Inference efficiency}
\label{sec:app-efficiency}

We additionally report the average end-to-end inference time per question and
peak GPU-memory usage on V$^*$Bench. Table~\ref{tab:app-efficiency} compares the
base model with representative training-free visual-search and
thinking-with-images methods under our evaluation setup. All methods are measured
under the same NVIDIA A800 hardware setup, and the reported latency is averaged
over all benchmark questions.

\begin{table}[t]
\centering
\small
\caption{Average inference efficiency on V$^*$Bench. Time is reported in
seconds per question and memory denotes peak GPU-memory usage. ViCrop$^\ast$
is our memory-optimized implementation of ViCrop. The original ViCrop
implementation runs out of memory even on an 80~GB GPU under this setup.}
\label{tab:app-efficiency}
\begin{tabular}{lcc}
\toprule
Method & Avg. time (s/question) & Peak GPU memory \\
\midrule
Base & 2.20 & -- \\
HiDe & 5.36 & 22~GB \\
DeepEyes & 12.56 & 21~GB \\
ViCrop & -- & OOM on 80~GB \\
ViCrop$^\ast$ & 5.98 & 21~GB \\
EviSpec & 6.43 & 22~GB \\

\bottomrule
\end{tabular}
\end{table}

EviSpec averages 6.43~s per question with a 22~GB peak memory footprint. Its
memory use matches HiDe and remains close to the optimized ViCrop$^\ast$ and
DeepEyes implementations. In latency, EviSpec incurs only a modest additional
cost over HiDe and ViCrop$^\ast$, while remaining substantially faster than
DeepEyes. The original ViCrop implementation does not fit within 80~GB of GPU
memory; ViCrop$^\ast$ therefore denotes our memory-optimized implementation,
which reduces peak usage to 21~GB and enables the reported 5.98~s/question
measurement.

\section{Supplementary Method and Proofs}
\label{sec:supplement}

This appendix expands the operators summarized in the main Method and proves the properties that follow from their construction. The results below are representational and algorithmic guarantees. They concern deterministic properties of the compiler and evidence operators; semantic relevance and answer accuracy are assessed by the matched experiments above.

The compiler and evidence-packing operators are shared by both localization
wrappers evaluated above. For completeness, the equations below spell out the
adaptive connected-region wrapper, which has the richer region operator; the
fixed-threshold wrapper replaces only the thresholding and region-formation
step. The request compiler, field separation, evidence assembly and final answer
contract are otherwise identical. The formal compiler and packing guarantees
therefore apply to both wrappers, while the expansion guarantee applies when
adaptive expansion is enabled.

\subsection{Notation and Complete Operators}
\label{sec:app-complete-operators}
Let $F_\theta$ denote the frozen MLLM used by the main method. Superscripts
$\mathrm{comp}$, $\mathrm{loc}$ and $\mathrm{ans}$ identify calls with different
fixed prompts but shared parameters $\theta$. Let $Q=(q,\mathcal{O})$ denote a
multiple-choice query, let $\bar Q=\operatorname{Serialize}(q,\mathcal{O})$ be
its complete serialized input string, and let $q_0$ be the prefix of $\bar Q$
before the first option marker. Define $s^T=\operatorname{Role}(\bar Q)$,
$s^C=\operatorname{Can}(\bar Q)$ and
$s^S=\operatorname{Can}(q_0)$. Here $\operatorname{Role}$ and
$\operatorname{Can}$ are calls to $F_\theta^{\mathrm{comp}}$ with the fixed
role-separated and canonical-form literals reproduced above; the deterministic
parser retains only the required tagged field and maps malformed or empty
fields to an empty request. The retained compiler sequence is
\begin{equation}
\mathbf{s}(\bar Q)=
\operatorname{UniqueText}(\widetilde{\mathbf{s}}(\bar Q)).
\label{eq:app-compiler}
\end{equation}
$\widetilde{\mathbf{s}}(\bar Q)=[s^T,s^C,s^S]$ is the ordered sequence before
filtering.
$\operatorname{UniqueText}$ removes empty requests and exact duplicates while
preserving the first occurrence. Each retained request is represented by
$s^r=(u_{r,1},\ldots,u_{r,K_r};m_r)$, where $u_{r,j}$ is a parsed entity field
and $m_r$ is an optional direct/relation tag. After tokenization,
$\mathcal{T}_{r,j}$ contains the signal tokens assigned to field $u_{r,j}$.

Let $\Omega_I$ be the source-pixel domain and
$\Omega_V=\{1,\ldots,H_V\}\times\{1,\ldots,W_V\}$ the original-image
visual-token grid. For a scalar map $X:\Omega_V\rightarrow\mathbb{R}$, define
\begin{equation}
\operatorname{range}_{\Omega_V}(X)=
\max_{p\in\Omega_V}X(p)-\min_{p\in\Omega_V}X(p),
\label{eq:app-range}
\end{equation}
and the sample-local normalization
\begin{equation}
\operatorname{N}(X)=
\frac{X-\min_{\Omega_V}X}{\operatorname{range}_{\Omega_V}(X)}.
\label{eq:app-normalization}
\end{equation}
Equation~\eqref{eq:app-normalization} applies when
$\operatorname{range}_{\Omega_V}(X)>0$; for a constant map, we define
$\operatorname{N}(X)\equiv0$. Thus every non-constant map is normalized
independently on its own visual grid, with per-question normalization for each
dataset.

For request $r$, let $\mathcal{T}_{r,j}$ be the token group assigned to its
$j$th parsed entity. The calibrated response of each signal token is computed
independently. At the fixed localization layer $\ell^\star$, the all-head
response is
\begin{equation}
A_{r,t}=
\frac{1}{|\mathcal{H}_F|}
\sum_{h\in\mathcal{H}_F}
A_{r,t}^{(\ell^\star,h)}.
\label{eq:app-attention}
\end{equation}
Its normalized form is
\begin{equation}
\bar A_{r,t}=\operatorname{N}(A_{r,t}).
\label{eq:app-normalized-attention}
\end{equation}
The instruction-prefix prior is
\begin{equation}
B_r(p)=\operatorname{median}_{u\in\mathcal{P}_r}\bar A_{r,u}(p).
\label{eq:app-prompt-prior}
\end{equation}
The nonnegative residual is
\begin{equation}
D_{r,t}(p)=[\bar A_{r,t}(p)-B_r(p)]_+.
\label{eq:app-token-response}
\end{equation}
Here $\mathcal{P}_r$ contains only the fixed instruction-prefix tokens. The
normalized residual is
\begin{equation}
R_{r,t}=\operatorname{N}(D_{r,t}).
\label{eq:app-entity-map}
\end{equation}
When $D_{r,t}\equiv0$, the implementation instead sets
$R_{r,t}=\bar A_{r,t}$. This direct fallback uses the normalized token map. The
implementation retains separate token maps through thresholding;
$\mathcal{T}_{r,j}$ is used later to group the independently
localized token boxes.

For each signal-token map, Yen's operator returns a threshold within the
observed support:
\begin{equation}
\tau_{r,t}=\operatorname{Yen}(R_{r,t}).
\label{eq:app-threshold}
\end{equation}
The associated high-response mask is
\begin{equation}
\mathcal{M}_{r,t}^{+}
=\{p\in\Omega_V:R_{r,t}(p)\geq\tau_{r,t}\}.
\label{eq:app-high-mask}
\end{equation}
Its eight-connected components are
\begin{equation}
\mathcal{C}_{r,t}^{+}=\operatorname{CC}_8(\mathcal{M}_{r,t}^{+}).
\label{eq:app-components}
\end{equation}
The maximum-mass component supplies the seed:
\begin{equation}
C_{r,t}^{*}
\in\operatorname*{arg\,max}_{C\in\mathcal{C}_{r,t}^{+}}
\sum_{p\in C}R_{r,t}(p).
\label{eq:app-seed}
\end{equation}
A fixed row-major tie rule makes the latter selection unique when several components have equal mass. Define the request-structural expansion flag
\begin{equation}
e_r=
\begin{cases}
1, & K_r>1,\\
0, & K_r=1.
\end{cases}
\label{eq:app-expansion-flag}
\end{equation}
Under the compiler contracts above, relation-mode $T$ requests contain separate
subject and reference fields, while multi-entity $C/S$ requests contain multiple
comma-delimited entity fields. This rule matches the operational behavior of the
original branch conditions, with parsed request structure determining expansion
across the $T/C/S$ labels.
If $e_r=0$, the selected token box is $b_{r,t}=b(C_{r,t}^{*})$. If $e_r=1$, let
\begin{equation}
    V_{r,t}^{-}
    =\{R_{r,t}(p):R_{r,t}(p)<\tau_{r,t}\}.
    \label{eq:app-low-support}
\end{equation}
If $V_{r,t}^{-}$ is nonempty, define
$\tau_{r,t}^{-}=\operatorname{Yen}(V_{r,t}^{-})$ and
\begin{equation}
\mathcal{M}_{r,t}^{-}
=\{p\in\Omega_V:R_{r,t}(p)\geq\tau_{r,t}^{-}\}.
\label{eq:app-low-mask}
\end{equation}
The expanded component is
\begin{equation}
C_{r,t}^{-}
=\operatorname{CC}_8(\mathcal{M}_{r,t}^{-};C_{r,t}^{*}).
\label{eq:app-expansion}
\end{equation}
The selected token box is $b_{r,t}=b(C_{r,t}^{-})$. The notation
$\operatorname{CC}_8(S;C^*)$ means the component of $S$ that contains seed
$C^*$. The containment proof below applies whenever $V_{r,t}^{-}$ is
nonempty. In the degenerate empty-support case, the implementation keeps the
high-threshold seed.

Each selected token box is mapped from $\Omega_V$ to normalized coordinates on
$\Omega_I$ and assigned to its parser field. Let
$\mathcal{B}_{r,j}^{0}$ be the boxes assigned to field $j$, and let
$h_{r,j}=b(\bigcup_{\beta\in\mathcal{B}_{r,j}^{0}}\beta)$ be their enclosing
rectangle. Field $j$ is fused only when $h_{r,j}$ is disjoint from every box
assigned to another field; otherwise its boxes remain separate. All guards are
evaluated against the pre-fusion boxes, so the output is independent of field
processing order. The final sequence $\mathcal{B}_r$ concatenates fields in
parser order and applies the fixed row-major tie rule within each field.

The evidence view is
\begin{equation}
    E_r=\operatorname{Assemble}\!\left(
    \operatorname{sort}_{\mathrm{row\mbox{-}major}}
    [\Pi(I,b):b\in\mathcal{B}_r]\right),
    \label{eq:app-view}
\end{equation}
where $\Pi(I,b)$ copies source pixels inside $b$ without resizing.
$\operatorname{Assemble}$ removes only horizontal and vertical source intervals
that intersect no retained box, using monotone source-to-canvas coordinate
maps. Overlapping source pixels are written once; source order and shared
geometry are therefore preserved while empty gaps become transparent. Empty
request branches are omitted. $\operatorname{UniqueView}$ removes a view only
when its shape, alpha support and visible pixel values exactly equal an earlier
view, retaining the first occurrence in $T,C,S$ order. Finally,
$\operatorname{InputPack}$ places the unmodified source image first, appends
the surviving evidence views in that order and supplies the unchanged question
to the answer call:
\begin{equation}
\mathbf{E}=\operatorname{UniqueView}
([E_r:r\in(T,C,S),\ E_r\ \text{defined}]).
\label{eq:app-evidence-sequence}
\end{equation}
\begin{equation}
\hat a=F_\theta^{\mathrm{ans}}
\bigl(\operatorname{InputPack}([I]\Vert\mathbf{E}),Q\bigr).
\label{eq:app-complete}
\end{equation}

\subsection{Complete Adaptive-Wrapper Inference Procedure}

For clarity, the full inference procedure is listed in execution order.

\begin{enumerate}
    \item Represent the multiple-choice query as $Q=(q,\mathcal{O})$, serialize it as $\bar Q=\operatorname{Serialize}(q,\mathcal{O})$, and split $\bar Q$ at the first standard option marker to obtain the option-free prefix $q_0$.
    \item Apply the role-separated and canonical-form output instructions to $\bar Q$, and apply the same canonical-form instruction to $q_0$, using the three constructions defined in the main paper; remove exact textual duplicates.
    \item For each remaining specification, insert it into the shared grounding prompt and run the frozen MLLM while recording the prescribed text-to-vision attention maps.
    \item Aggregate and normalize the maps, estimate the median prefix prior, and compute each calibrated signal-token map in Eqs.~\eqref{eq:app-attention}--\eqref{eq:app-entity-map}.
    \item Apply the sample-local Yen threshold to each token map and select its maximum-mass connected seed. Apply the lower-threshold connected expansion according to Eq.~\eqref{eq:app-expansion-flag} whenever the parsed request contains more than one entity field ($K_r>1$).
    \item Map the selected grid boxes back to the source image, apply the field-guarded fusion rule, and assemble one evidence view per request construction.
    \item Remove duplicate evidence views, append the remaining views to the original image, restore the unchanged multiple-choice question, and decode the final option with the frozen MLLM.
\end{enumerate}

The procedure uses frozen model weights and fixed compiler, localization and
answering operators. Model-specific layers and image budgets are configured once
for each model family and reported with the experimental setup.

\subsection{Question-Compilation Guarantees}

\paragraph{Proposition 1 (option-order invariance of the isolated branch).}
Let $Q=(q,\mathcal{O})$ and $Q'=(q,\pi(\mathcal{O}))$ differ only by a permutation $\pi$ of text after the first option marker. Then
\begin{equation}
    s^{S}(Q)=s^{S}(Q'),
    \qquad
    G_{S}(I,Q;F)=G_{S}(I,Q';F),
    \label{eq:app-option-invariance}
\end{equation}
provided that tokenization, frozen-model inference and the region operators are run with the same state and tie rules.

\emph{Proof.}
Write $P(Q)$ for the prefix of the serialized question ending immediately before the first option marker. By assumption, $Q$ and $Q'$ have the same text before this marker, and the permutation acts only on the suffix. Hence $P(Q)=P(Q')$. The option-isolated compiler is the composition $\operatorname{Can}\circ P$, so
\[
    s^{S}(Q)
    =\operatorname{Can}(P(Q))
    =\operatorname{Can}(P(Q'))
    =s^{S}(Q').
\]
The resulting grounding prompt and token sequence are identical. With the same
image and frozen model, the recorded attention maps are identical.
Equations~\eqref{eq:app-attention}--\eqref{eq:app-view} therefore produce the
same regions and evidence view for the option-isolated branch. The final answer
stage receives the complete option sequence. $\square$

\paragraph{Proposition 2 (separate grounding of successfully extracted fields).}
Suppose the role-separated full-query parser returns an ordered, comma-delimited list $s^{T}=[u_1,\ldots,u_M]$. Before evidence-view fusion, EviSpec assigns every $u_i$ a distinct token group $\mathcal{T}_{T,i}$ and constructs its token maps independently by Eq.~\eqref{eq:app-entity-map}.

\emph{Proof.}
The comma delimiter defines $M$ disjoint fields. Tokenization assigns every
token position to one field, so for $i\neq j$,
$\mathcal{T}_{T,i}\cap\mathcal{T}_{T,j}=\varnothing$. Equation~\eqref{eq:app-entity-map}
constructs every signal-token map independently. Field interactions occur later
during guarded box fusion and view construction. $\square$

\subsection{Calibration Guarantees}

\paragraph{Proposition 3 (bounded calibrated response).}
For every specification $r$, token $t$ and pixel $p$,
\begin{equation}
    0\leq \bar A_{r,t}(p)\leq 1,\qquad
    0\leq R_{r,t}(p)\leq 1.
    \label{eq:app-bounds}
\end{equation}

\emph{Proof.}
If $A_{r,t}$ is non-constant, Eq.~\eqref{eq:app-normalization} subtracts its minimum and divides by its positive range. The minimum is mapped to zero, the maximum to one, and every intermediate value lies between them. If the map is constant, the definition returns zero, so $\bar A_{r,t}\in[0,1]$ in both cases.

The positive-part operator makes $D_{r,t}$ nonnegative. When its range is
positive, the same normalization argument gives
$\operatorname{N}(D_{r,t})\in[0,1]$. A nonzero constant residual is mapped to
zero by the constant-map rule. When $D_{r,t}\equiv0$, the fallback defined
after Eq.~\eqref{eq:app-entity-map} uses $\bar A_{r,t}$, which is already
bounded. Thus $R_{r,t}\in[0,1]$. $\square$

\paragraph{Proposition 4 (median robustness to minority prefix contamination).}
Fix a pixel $p$ and suppose the prefix contains $m$ responses. If fewer than $m/2$ responses are changed arbitrarily while every unchanged response lies in $[a,b]$, then the pointwise median prior remains in $[a,b]$.

\emph{Proof.}
Let $k<m/2$ responses be arbitrary and let the remaining $m-k>m/2$
responses lie in $[a,b]$. More than half of all values are at least $a$ and at
most $b$, so their median lies in $[a,b]$. Therefore
$a\leq B_r(p)\leq b$. $\square$

\paragraph{Proposition 5 (cancellation of a shared prompt response).}
Assume that, after token-map normalization, a spatial response $\delta:\Omega\rightarrow\mathbb{R}$ is added to every entity-token map and every prefix-token map:
\[
    \bar A'_{r,t}=\bar A_{r,t}+\delta,\qquad
    \bar A'_{r,u}=\bar A_{r,u}+\delta
    \quad\text{for all }u\in\mathcal{P}_r.
\]
Then the pre-normalized calibrated residual is unchanged:
\begin{equation}
    [\bar A'_{r,t}-\operatorname{median}_{u\in\mathcal{P}_r}\bar A'_{r,u}]_+
    =
    [\bar A_{r,t}-\operatorname{median}_{u\in\mathcal{P}_r}\bar A_{r,u}]_+ .
    \label{eq:app-cancellation}
\end{equation}

\emph{Proof.}
At each pixel $p$, adding the same scalar $\delta(p)$ to every member of a finite set shifts every ordered value by $\delta(p)$. In particular,
\[
 \operatorname{median}_{u}(\bar A_{r,u}(p)+\delta(p))
 =
 \operatorname{median}_{u}\bar A_{r,u}(p)+\delta(p).
\]
Substituting this identity gives
\[
 \bar A_{r,t}(p)+\delta(p)
 -\operatorname{median}_{u}\bar A_{r,u}(p)-\delta(p)
 =
 \bar A_{r,t}(p)-\operatorname{median}_{u}\bar A_{r,u}(p).
\]
The two arguments of the positive-part operator are identical at every pixel,
which proves Eq.~\eqref{eq:app-cancellation}. Subsequent normalization is applied
to the same residual map. $\square$

\subsection{Region-Formation Guarantees}

\paragraph{Proposition 6 (mass-optimal seed within the threshold mask).}
For a fixed signal-token map and threshold mask, the selected component satisfies
\begin{equation}
    \sum_{p\in C_{r,t}^{*}}R_{r,t}(p)
    \geq
    \sum_{p\in C}R_{r,t}(p),
    \qquad
    \forall C\in\mathcal{C}_{r,t}^{+}.
    \label{eq:app-seed-optimal}
\end{equation}

\emph{Proof.}
The visual grid contains finitely many eight-connected components.
Equation~\eqref{eq:app-seed} selects $C_{r,t}^{*}$ by maximizing the displayed
mass function. A fixed row-major rule resolves ties while preserving the
inequality. $\square$

\paragraph{Proposition 7 (enabled expansion contains the seed).}
Assume $V_{r,t}^{-}$ in Eq.~\eqref{eq:app-low-support} is nonempty and the threshold returned by Yen's operator lies within its input support. Then
\begin{equation}
    C_{r,t}^{*}\subseteq C_{r,t}^{-},
    \qquad
    b(C_{r,t}^{*})\subseteq b(C_{r,t}^{-}).
    \label{eq:app-containment}
\end{equation}

\emph{Proof.}
Every value in $V_{r,t}^{-}$ is strictly smaller than $\tau_{r,t}$. A threshold chosen within that support therefore satisfies $\tau_{r,t}^{-}<\tau_{r,t}$. Every seed pixel belongs to the high mask, so for $p\in C_{r,t}^{*}$,
\[
    R_{r,t}(p)\geq\tau_{r,t}>\tau_{r,t}^{-}.
\]
Hence every seed pixel also belongs to $\mathcal{M}_{r,t}^{-}$. The operator $\operatorname{CC}_8(\mathcal{M}_{r,t}^{-};C_{r,t}^{*})$ selects the lower-mask component containing those pixels, which proves $C_{r,t}^{*}\subseteq C_{r,t}^{-}$. The axis-aligned bounding rectangle of a set cannot shrink when the set is enlarged: each coordinate minimum can only decrease or remain unchanged, and each maximum can only increase or remain unchanged. Therefore the bounding-box inclusion follows. $\square$

\paragraph{Proposition 8 (parser-field separation invariant of guarded fusion).}
For every field $j$ accepted for fusion, its enclosing rectangle is disjoint
from every pre-fusion box assigned to a field $k\neq j$.

\emph{Proof.}
Let $h_{r,j}$ be the enclosing rectangle of the boxes in field $j$. By the
definition of guarded fusion, $\mathcal{B}_{r,j}^{0}$ is replaced by
$\{h_{r,j}\}$ only if
$h_{r,j}\cap b=\varnothing$ for every
$b\in\mathcal{B}_{r,k}^{0}$ and every $k\neq j$. Thus every accepted fused
rectangle satisfies the stated separation property directly. Because all
guards use the same pre-fusion box sets, the decision is independent of the
order in which fields are inspected. $\square$

\subsection{Evidence-View Guarantees}

\paragraph{Proposition 9 (source support of visible evidence and order preservation).}
Let $\operatorname{supp}_{\alpha}(E_r)$ be the pixels with nonzero alpha in the packed evidence view. Every visible evidence pixel is copied from a selected source-image box:
\begin{equation}
    \forall p\in\operatorname{supp}_{\alpha}(E_r),\quad
    \exists b\in\mathcal{B}_r,\ z\in b
    \ \,\text{such that}\ \, E_r(p)=I(z).
    \label{eq:app-pixels}
\end{equation}
If two non-overlapping boxes satisfy $b_1\prec_{\mathrm{rm}}b_2$, their extracted contents appear in that order in $E_r$.

\emph{Proof.}
By Eq.~\eqref{eq:app-view}, $\operatorname{Assemble}$ receives the crops
$\Pi(I,b)$ for $b\in\mathcal{B}_r$ and copies each retained source region to its
compressed coordinate interval. Every visible output pixel therefore equals a
source pixel inside a selected box, which proves Eq.~\eqref{eq:app-pixels}.

For order preservation, the assembly operator sorts the source $x$- and
$y$-boundaries and maps them to compressed coordinates with monotone coordinate
maps. The row and column precedence defined by $\prec_{\mathrm{rm}}$ is preserved.
The complete input in Eq.~\eqref{eq:app-complete} also retains the original image
for global context. $\square$

\subsection{Computational Complexity}

Let $R=|\mathbf{s}(\bar Q)|$, let $T=\sum_{r,j}|\mathcal{T}_{r,j}|$ be the total number of grounding tokens, let $N=|\Omega_V|=H_VW_V$, and let $H_a=|\mathcal{H}_F|$ be the number of aggregated attention heads at the fixed localization layer. Attention aggregation and calibration require
\begin{equation}
    \mathcal{O}(T H_a N)
    \quad\text{time and}\quad
    \mathcal{O}(N+H_aN)
    \quad\text{working memory when token maps are streamed}.
    \label{eq:app-attn-complexity}
\end{equation}
Thresholding and connected-component extraction are linear in $N$ for each token map. Sorting $|\mathcal{B}_r|$ box boundaries for view construction costs $\mathcal{O}(|\mathcal{B}_r|\log|\mathcal{B}_r|)$. The complete evidence stage is therefore bounded by
\begin{equation}
    \mathcal{O}\!\left(
    T H_a N+TN+
    \sum_{r=1}^{R}|\mathcal{B}_r|\log|\mathcal{B}_r|
    \right),
    \label{eq:app-total-complexity}
\end{equation}
in addition to the frozen-model forward passes and final generation.

The propositions establish deterministic properties of question compilation,
prefix calibration, region expansion, guarded fusion and evidence-view assembly.
The matched prompt-family comparisons evaluate request construction, and the
exact-geometry experiment evaluates the contribution of selected visual content.

\end{document}